\documentclass{article}

\usepackage[preprint]{neurips_2025}
\usepackage{mathtools}
\usepackage{microtype}
\usepackage{graphicx}
\usepackage{subfigure}
\usepackage{booktabs} % for professional tables
\usepackage{subcaption}
\usepackage{float}
\usepackage{scrextend}
\usepackage{amsthm}
\usepackage{amsmath} 

\usepackage[utf8]{inputenc} % allow utf-8 input
\usepackage{longtable}

\usepackage{hyperref}
\usepackage{pbox}

\newcommand{\ignore}[1]{}

\usepackage{amsmath}
\usepackage{amssymb}
\usepackage{mathtools}
\usepackage{amsthm}
\usepackage{caption}

\theoremstyle{plain}
\newtheorem{theorem}{Theorem}[section]

\theoremstyle{definition}

\theoremstyle{remark}

\usepackage[T1]{fontenc}    % use 8-bit T1 fonts
\usepackage{hyperref}       % hyperlinks
\usepackage{url}            % simple URL typesetting
\usepackage{booktabs}       % professional-quality tables
\usepackage{amsfonts}       % blackboard math symbols
\usepackage{nicefrac}       % compact symbols for 1/2, etc.
\usepackage{microtype}      % microtypography
\usepackage{xcolor}         % colors

\usepackage{siunitx}
\usepackage{booktabs}
\usepackage{enumitem}
\newtheorem*{theorem*}{Theorem}
\title{Learning Representational Disparities}

\author{%
  Pavan Ravishankar$^{1,2}$, \textbf{Rushabh Shah}$^{1,2}$, \textbf{Daniel B. Neill}$^{1,2,3,4}$ \\
  Machine Learning for Good Laboratory, New York University (NYU)$^{1}$\\
  Courant Institute of Mathematical Sciences, Department of Computer Science, NYU$^{2}$\\
  Robert F. Wagner Graduate School of Public Service, NYU$^{3}$ \\ Center for Urban Science and Progress, Tandon School of Engineering, NYU$^{4}$ \\ {\{pr2248@nyu.edu, rushabh.shah@nyu.edu, daniel.neill@nyu.edu\}}
}

\begin{document}
\maketitle

\begin{abstract}
We propose a fair machine learning algorithm to model interpretable differences between observed and desired human decision-making, with the latter aimed at reducing disparity in a downstream outcome impacted by the human decision. Prior work learns fair representations without considering the outcome in the decision-making process. We model the outcome disparities as arising due to the different representations of the input seen by the observed and desired decision-maker, which we term representational disparities. Our goal is to learn interpretable representational disparities which could potentially be corrected by specific nudges to the human decision, mitigating disparities in the downstream outcome; we frame this as a multi-objective optimization problem using a neural network. Under reasonable simplifying assumptions, we prove that our neural network model of the representational disparity learns interpretable weights that fully mitigate the outcome disparity. We validate objectives and interpret results using real-world German Credit, Adult, and Heritage Health datasets.
\end{abstract}

\section{Introduction}
Many human decisions are made with the aid of machine learning algorithms, which have had a significant impact across various domains such as healthcare \cite{de2018clinically}. Machine learning algorithms have frequently been criticized in the fairness literature due to their potential to create or exacerbate disparities~\cite{angwin2016machine,barocas2017fairness,mehrabi2021survey}.  However, human decisions can also exhibit demographic and other biases, whether intentional (e.g., racial animus) or unintentional, with undesirable impacts, e.g., in policing~\cite{Gelman01092007}.  Moreover, even when presented with ostensibly ``fair'' algorithmic predictions, human decision-makers tend to deviate from these predictions in systematically biased ways~\cite{green2020algorithm}, suggesting a need to provide more concrete and interpretable recommendations for behavioral change.

%An important caveat in the criticism, often incorrectly presumed even in machine learning communities, is that the algorithm is the sole decision maker disregarding the role of a human in the decision-making process. One way to address the underlying concern is to design algorithms with an \textit{algorithm-in-the-loop} framework \cite{green2020algorithm}, which privileges humans and uses algorithms as an aid to improve human decision-making. Thus, the onus of equity lies on the human, not the algorithm. 

In this paper, we propose an algorithmic aid to mitigate downstream disparity in outcomes resulting from human decisions. We describe a methodological solution to model and correct differences between an \emph{observed} human decision-making process, with resulting disparity in outcomes impacting some protected class, and a \emph{desired} (fairer) human decision-making process, which differs from the observed decisions in a systematic and interpretable way, and 
mitigates the observed disparity. The differences serve as concrete aids to nudge the human decision-maker toward fairer behavior~\cite{thaler2009nudge}, as measured by a reduction in outcome disparity.  This approach aligns with the core principles of the \emph{algorithm-in-the-loop framework}~\cite{green2020algorithm}, preserving the agency of the human decision-maker while providing them with algorithmic recommendations geared toward improving the fairness of the combined (algorithm + human) system.

%Moreover, such an algorithm gives leeway to the human decision-maker to consider the sociotechnical context while changing their behavior. Providing leeway underscores the intent of the algorithm to make the human privileged to make final decisions. By doing so, the algorithm aligns with the core principles of the algorithm-in-the-loop framework, which emphasizes the development of systems for integration into sociotechnical contexts rather than isolated decision-making \cite{green2020algorithm}. 

\begin{figure}[t]
    \centering
    \includegraphics[width=2.5cm,height=2.5cm]{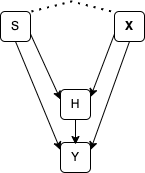}
    \caption{Data Generation Process}
    \label{datagengraph}
\end{figure}

Concretely, we assume a decision process (Figure~\ref{datagengraph}) in which a human decision-maker (e.g., housing agency) makes a consequential decision $H$ (e.g., whether to give an applicant a housing voucher) that impacts a downstream outcome $Y$ (e.g., whether the applicant is able to obtain housing).  The decision could be based on the applicant's values of a binary sensitive attribute $S$ and other non-sensitive attributes $\mathbf{X}$. Critically, the human decision $\mbox{Pr}(H\:|\:S,\mathbf{X})$, the outcome model $\mbox{Pr}(Y\:|\:H,S,\mathbf{X})$, or both, could be biased in a way that differentially impacts the protected class $S=1$.  For example, a biased human decision-maker may provide housing vouchers less often to minoritized individuals, or such individuals may be less likely to obtain housing with or without a voucher. In the former case, the desired decision may be to correct $H$ to be uncorrelated with $S$, but in the latter case, outcome disparities may persist unless minoritized individuals receive housing vouchers \emph{more} often than non-minoritized individuals to compensate for the downstream biases in the outcome variable.

%Suppose the observed decision-maker uses the racial attribute to allocate housing vouchers, resulting in downstream disparity towards low-income racial groups and a fair (desired) decision-maker does not use the racial attribute to allocate more housing vouchers to deprived communities, reducing downstream disparity towards low-income racial groups. Then, the former could be nudged not to use the racial attribute \cite{thaler2009nudge}. 

In this paper, based on the observation that value-based (economic) decisions generally rely on memory-based representations \cite{smith2021mental, eichenbaum2014can, squire1991medial, suzuki2009memory},
we model the differences between the observed and desired decision-making processes, and resulting disparities in outcomes, as arising due to different \emph{representations} of the inputs seen by the observed and desired decision-makers, which we term \emph{representational disparities}. Concretely, we learn a shallow, mechanistically interpretable neural network model where a portion of the hidden layer models the representational disparities, thus explaining the differences between the observed and desired decisions and their resulting outcomes.

%We propose an algorithm, aligned with the \textit{ algorithm-in-the-loop} framework, to model disparities between the observed decision-maker and the fair (desired) decision-maker. 

The key contributions of the paper are as follows:
\begin{enumerate}
\item \textit{Methodology:} We mathematically formulate the goal of the desired human decision-maker to reduce the disparity in outcomes between the protected and non-protected class.
\item \textit{Optimization:} We learn the representational disparities between the observed and the desired decision-makers by formulating the problem as an multi-objective optimization problem with the primary objectives of 
(i) mitigating outcome disparity between protected and non-protected class using fair (desired) decisions; and (ii) learning representational disparities that represent \emph{interpretable} differences between observed and desired decisions.
\item \textit{Theory:} Under reasonable simplifying assumptions, we prove that the weights learned by the optimization procedure result in representational disparities that are interpretable
 and if corrected, will fully mitigate the disparity in outcomes.  In more general settings, convergence to these desired weights is not guaranteed but can be achieved in practice via multiple random initializations.
\item \textit{Experiments:} We validate the optimization objectives using synthetically created data sets and investigate representational disparities using real-world German Credit, Adult Income, and Heritage Health datasets. We compare our results to a foundational work \emph{Fair Representation Learning} \cite{zemel2013learning},
and show that our methodology provides multiple advantages including higher accuracy, increased interpretability and consistency in its recommendations (thus facilitating nudges toward fairness), and greater reduction in outcome disparity, by accounting for biases in the distribution of the outcome conditional on the human decision.

%We argue that our methodology performs similarly in all metrics reported in \cite{zemel2013learning}, and additionally provides two benefits: (1)Interpretability in which the attribute(s) primarily responsible for the reduction in disparity is identified. (2)Increased reduction in disparity in an often ignored, yet important, setting in which outcome and human decisions differ (details in Section 5).

\end{enumerate}
The remainder of the paper is organized as follows: Section 2 reviews related work. Section 3 describes the model by formalizing the notations, setting, and optimization problem. Section 4 discusses theorems that prove that the representational disparities learned are interpretable and mitigate disparities in outcomes. Section 5 validates each objective of the optimization problem using synthetic data, investigates disparities in real-world datasets, and compares our model with competing approaches. Finally, Section 6 discusses conclusions and future work. 

\section{Related Work}
A vast corpus of literature on algorithmic fairness mainly considers the algorithm, disregarding the role of the human. Stage-specific fairness approaches analyze fairness from the lens of a particular stage of the machine learning pipeline such as data or predictions~\cite{mehrabi2021survey, barocas2017fairness}, while pipeline-aware approaches analyze how bias propagates across multiple pipeline stages such as from data to the predictions, either qualitatively \cite{suresh2021framework,black2023toward} or quantitatively \cite{ravishankar2023provable}. Only a few works have discussed the fairness problem with both the algorithm and the human decision-maker in purview: our work is built on the \textit{algorithm-in-the-loop} framework \cite{green2020algorithm}, which assumes that a human decision-maker makes a consequential decision with support from an algorithmic prediction or recommendation. 

Numerous works have proposed methodologies and frameworks to learn fair representations \cite{zemel2013learning, zhao2022inherent, zhao2019conditional, madras2018learning, luo2023learning}. A framework for fair representation is proposed by defining an $\epsilon$-fair representation model and a representation algorithm with high-confidence fairness guarantees \cite{luo2023learning}. With $\epsilon$-fair representation model defined as the disparity not exceeding $\epsilon$ for every model, an upper bound for the disparity, formulated as the mutual information between the input representation and the sensitive attribute, is used to prove high-confidence guarantees as it is model agnostic. Since computing mutual information is intractable, a tractable upper bound is formulated to prove confidence guarantees. While this work proposes theoretical definitions and guarantees of fair representation learning for generic tasks, our work provides a concrete methodological solution to a specific problem.

Within the applied literature, the foundational work of \cite{zemel2013learning} has proposed a methodology to learn fair representations with three primary objectives: to preserve utility, mitigate demographic disparity, and minimize cross-entropy. We compare our model with this approach on the German Credit, Adult, and Health datasets, demonstrating improvements in accuracy, interpretability, and reduced disparity. On a similar theme,~\cite{madras2018learning} uses adversarial learning to learn fair representations by minimizing classification loss, reconstruction loss, and disparity in fairness. It proves that minimizing the disparity in fairness is equivalent to maximizing adversarial loss. Our work differs in the methodology and our intent in learning fair representations: to elicit the differences between the desired and observed decision-maker so that these differences can be corrected and outcome disparities reduced. 

Critically, none of the previous approaches to learning fair representations can be easily generalized to the case where there is a outcome $Y$ downstream of the human decision $H$, and the distribution of $Y$ given $H$ may be biased against the protected class.  In this case, the goal is not simply to remove the impact of the sensitive attribute $S$ by making $H$ independent of $S$, but instead to modify $H \:|\: S$  so that it corrects the downstream disparity in $Y$.

\section{Our Model}
\label{ourmodel}
In this section, we formalize the model and methodology, which encompasses the data generation process, neural network architecture, and optimization process. 

\textbf{Notation and Data Generation Process}\\
Let $S$ be a binary random variable that denotes belonging to a sensitive demographic group (e.g., defined by race and/or gender), where $S=1$ represents the protected class.
$\mathbf{X}=\{X_1,...,X_n\}$ is a vector of random variables that denotes the attributes of the individual, excluding the sensitive attribute $S$ (e.g., a housing applicant's record excluding race/ethnicity). $H$ is a binary random variable that denotes a human decision such as the allocation of vouchers, and $Y$ is a binary random variable that denotes an outcome such as successful acquisition of housing. As shown in Figure \ref{datagengraph} above, $S$ and $\mathbf{X}$ are the inputs, $H$ is the human decision made using $S$ and $\mathbf{X}$, and $Y$ is the outcome decided using $S$, $\mathbf{X}$, and $H$. $S$ could be correlated with $\mathbf{X}$. We assume that $S$, $\mathbf{X}$, $H$, and $Y$ are observed. Let $T$ denote the training dataset with each point $\{S=s, \mathbf{X}=\mathbf{x}, H=h,Y=y\}$. Note that $Y$ depends only on $S$, $\mathbf{X}$, and $H$, and not on how $H$ is generated, that is, whether $H$ is generated by the fair (desired) or observed human decision maker.

\begin{figure}[t]
    \centering
    \includegraphics[width=12cm,height=3cm]{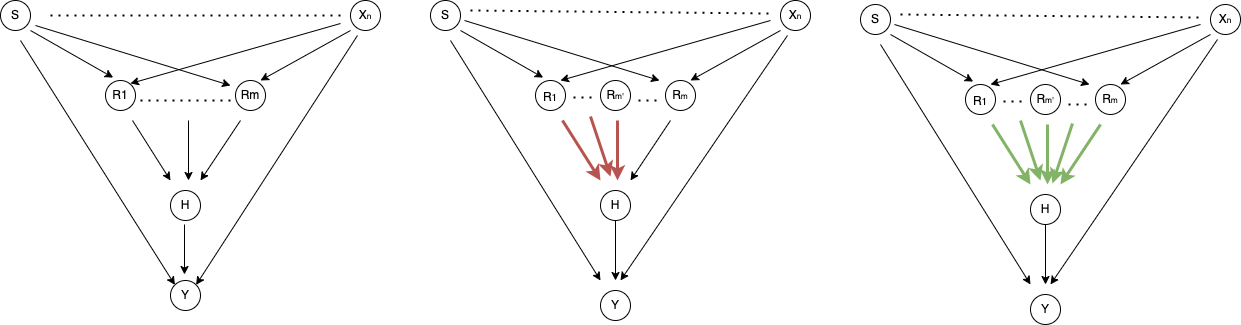}
    \caption{Architecture \textbf{(left)} with nodes used by the observed \textbf{(middle)} and desired human \textbf{(right)}}
    \label{neural_network}
\end{figure}

\textbf{Architecture, Decision Makers, and Representational Disparities}\\
As shown in Figure~\ref{neural_network}, we propose a shallow, mechanistically interpretable neural network architecture to simultaneously represent both observed and desired human decisions, with a portion of the hidden layer devoted to modeling representational disparities: the differences between observed and desired representations that explain the downstream outcome disparities.

The architecture is comprised of the following four layers: (1) Input layer, consisting of \{$S$, $\mathbf{X}$\}. (2) Internal representation of the input, $\mathbf{R}=\{R_1, ..., R_m\}$. The weights from the first layer to the second layer are denoted by $w_{ij}$, where $i$ is a node in the first layer, $i \in \{S, \mathbf{X}\}$, and $j$ is a node in the second layer, $j \in \{R_1, ..., R_m\}$. Each node $R_i$ has bias denoted as $\text{bias}_{R_i}$. (3) Human decision $H$ with a sigmoid activation function. The weights from the second layer to the third layer are denoted by $w_{i}$, where $i$ is a node in the second layer, $i \in \{R_1, ..., R_m\}$. (4) Outcome $Y$ with a sigmoid activation function with weights from \{$S$, $\mathbf{X}, H$\} to $Y$. A ReLU activation function aids in interpretability by identifying neurons that activate a node. 

The observed and desired decision-makers are assumed to differ in the internal representation of the input used to decide $H$. We assume that the former uses only a subset of the representation nodes, $R_1$ to $R_{m'}$, to decide $H$, while the latter uses all of the representation nodes $R_1$ to $R_{m}$ to decide $H$. Thus, nodes $R_{m'+1}$ to $R_{m}$ capture the representational disparities.

\textbf{Objectives}\\
The goal is to learn representational disparities by minimizing the four objectives described below. 

\textbf{Objective $A$:} The first objective is to mitigate outcome disparity between the protected class $S=1$ and non-protected class $S=0$ using the desired decision-maker, which we formulate as minimizing
\begin{align}
A &= \big|\text{Pr}(Y=1\:|\:S=1) - \text{Pr}(Y=1\:|\:S=0)\big| \label{initform} \\ &=\big|\underset{\mathbf{X=x}, H=h}{\sum}\text{Pr}(Y=1\:|\:\mathbf{X=x},S=1,H=h)\text{Pr}_{\text{des}}(H=h\:|\:\mathbf{X=x},S=1)\text{Pr}(\mathbf{X=x}\:|\:S=1) \nonumber \\ &~~-\underset{\mathbf{X=x}, H=h}{\sum}\text{Pr}(Y=1\:|\:\mathbf{X=x},S=0,H=h) \text{Pr}_{\text{des}}(H=h\:|\:\mathbf{X=x},S=0)\text{Pr}(\mathbf{X=x}\:|\:S=0) \big|, \nonumber
\end{align}
where $\text{Pr}_{\text{des}}$ represents the distribution of the desired decision-maker's decision $H$ (using all representation nodes $R_1$ to $R_m$) conditioned on inputs. This equation is obtained by factoring $\text{Pr}(S, \mathbf{X}, H, Y)$ according to the Bayesian network in Figure~\ref{datagengraph} \cite{koller2009probabilistic}. Note that demographic parity is a common fairness notion \cite{barocas2017fairness} and a reasonable formulation in value-based decisions such as house allocation.

\textbf{Objective $B$:} The second objective is to learn \emph{interpretable} representational disparities between the observed and the desired decision-maker, which we formulate as minimizing the sum of $L1$ regularization terms,
\begin{align}
B = &\underset{i \in \{m'+1,..., m\}}{\textstyle \sum} ||w_{\mathbf{X}R_i}||_1 + |w_{SR_i}| + |w_i| + |\text{bias}_{R_i}|, \label{obj2} \\ &\text{where}~||w_{\mathbf{X}R_i}||_1 = \underset{\substack{A \in \mathbf{X}}}{\textstyle \sum} |w_{AR_i}|. \nonumber
\end{align}
Note that all weights in Eq. \ref{obj2} are associated with the nodes that capture the representational disparity, that is, $R_{m'+1}, ...., R_{m}$. The first two terms penalize the incoming weights, the third term penalizes the outgoing weights, and the fourth term penalizes the bias term. We write $w$ instead of $w_i$ when only one representational node is used to capture the disparity. 
%https://medium.com/one-minute-machine-learning/l1-regularization-one-minute-summary-b407940e52f#:~:text=L1%20Regularization%20is%20a%20technique,a%20model%20more%20visually%20interpretable.
We employ $L1$ regularization to encourage sparsity, reducing model complexity by zeroing out less important weights, and thereby making the model interpretable \cite{goodfellow2016deep}. For instance, if there is no unfairness towards any sensitive group $S$, then $L1$ regularization encourages a zero weight from $S$ to all representational disparity nodes. More generally, this formulation encourages the model to learn a desired decision process that is similar to the observed decision process, with only those differences (represented by $R_{m'+1} \ldots R_m$) that are necessary to explain and mitigate outcome disparities.

\textbf{Objective $C$:}
The third objective is to correctly model the observed decision process, $\text{Pr}_{\text{obs}}(H=1\:|\:\mathbf{X=x}, S=s)$, which we formulate as minimizing the binary cross-entropy loss on training data,
\begin{align}
C = \frac{1}{|T|} \sum^{|T|}_{i=1}-h_i\ln \text{Pr}_{\text{obs}}(H=1 \:|\:\mathbf{x}_i, s_i)-(1-h_i)\ln \left(1-\text{Pr}_{\text{obs}}(H=1\:|\:\mathbf{x}_i,s_i)\right),
\end{align}
where $\text{Pr}_{\text{obs}}$ represents the distribution of the observed decision-maker's decision $H$ (using only representation nodes $R_1$ to $R_{m'}$) conditioned on inputs.
Since the desired decision-maker uses additional representation nodes along with the nodes used by the observed decision-maker, it is imperative to accurately learn the weights corresponding to the observed decision-maker to precisely interpret the representational disparity. Hence, a large weight is assigned to Objective $C$ compared to Objectives $A$ and $B$. 

\textbf{Objective $D$:} The fourth objective is to learn the outcome process, $\text{Pr}(Y=1 \:|\: \mathbf{X=x}, S=s)$, which we formulate as minimizing the binary cross-entropy loss,
\begin{align}
D = \frac{1}{|T|} \sum^{|T|}_{i=1}-y_i\ln \text{Pr}(Y=1 \:|\: \mathbf{x}_i, s_i, h_i)-(1-y_i)\ln \left(1-\text{Pr}(Y=1 \:|\: \mathbf{x}_i, s_i, h_i) \right).
\end{align}
We assume that the outcome process does not change given the inputs and the human decision. Hence, a large weight is assigned to Objective $D$ compared to Objectives $A$ and $B$ to learn the weights corresponding to the outcome process accurately. 

\textbf{Total Loss:} The objective is to minimize the total loss,
\begin{align}
aA + bB + cC + dD, \label{totalloss1}
\end{align}
where $c \gg a, c = d$, and $b = 1-a$ with $0 < a < 1$ to capture trade-off between Objectives $A$ and $B$.

\section{Theoretical Results}
\label{sec:theory}
In this section, we derive theoretical results to prove that the weights learned by the neural network are interpretable and mitigate the observed disparity. More precisely, we state three theorems, with proofs provided in Appendix~\ref{proofs}.  These theorems rely on the following three simplifying assumptions:\\[1ex]
\noindent (A1) The weights of the observed decision process $\text{Pr}_{\text{obs}}(H=1\:|\:\mathbf{X},S)$, and the weights of the outcome process $\text{Pr}(Y=1\:|\:\mathbf{X},S,H)$,
are learned from training data $T$ and fixed at these values.\\
\noindent (A2) The sensitive attribute $S$ is independent of the non-sensitive attributes $\mathbf{X}$. \\
\noindent (A3) The outcome $Y$ is conditionally independent of the sensitive attribute $S$ given $H$. % modified after main paper submission DBN

Assumption (A1) simplifies proofs by disregarding Objectives $C$ and $D$, but similar weights are learned without this assumption when $c$ and $d$ are much larger then $a$ and $b$. (A2) is made to simplify the proof, and realizable as attributes in $\mathbf{X}$ that are highly correlated with $S$ can be removed. (A3) also simplifies the proof, and is feasible as the outcome need not depend on $S$ to mitigate disparity.

\textbf{Theorem 4.1} considers a further simplified setting with three additional assumptions:\\[1ex] 
\noindent(A4) There are no non-sensitive attributes ($\mathbf{X}=\emptyset$).\\ 
\noindent(A5) There is only a single representational disparity node ($m=m'+1$), denoted as $R'$.\\
\noindent(A6) Disparity loss substantially outweighs interpretability loss ($a \approx 1$, $b \approx 0$).

Theorem 4.1 shows that with appropriate initialization of the network weights, the learned weights converge to the global minimum loss, with weights on the representational disparity node that are interpretable and fully mitigate the outcome disparity.

\begin{theorem}
Assume the data generating process and neural network architecture in Figures~\ref{datagengraph}-\ref{neural_network} and assumptions (A1)-(A6) above. Here the decision $H$ depends only on $S$, and the outcome $Y$ depends only on $H$. Let $\alpha = \text{Pr}(Y=1\:|\: H=1)-\mbox{Pr}(Y=1\:|\:H=0)$, $\alpha \ne 0$.  Moreover, assume that there is $\delta$-unfairness towards $S=1$ in the observed decision $H$, $\delta = \text{logit}(H=1\:|\: S=1)-\mbox{logit}(H=1\:|\:S=0)$, $\delta \ne 0$. Suppose the desired decision $D_{\mathbf{w}}(s) = \text{Pr}_\text{des}(H=1 \:|\: S=s)$, parameterized by weight vector $\mathbf{w} = (w, w_{SR'},\text{bias}_{R'})$, is learned using training data $T$ by minimizing the total loss $L_{\mathbf{w}}$, where
\begin{align*}
&L_{\mathbf{w}} = aA_{\mathbf{w}} + bB_{\mathbf{w}}, \\
&A_{\mathbf{w}} = |\alpha||D_{\mathbf{w}}(1)-D_{\mathbf{w}}(0)|, \text{ and} \\  &B_{\mathbf{w}} = |w| + |w_{SR'}| + |\text{bias}_{R'}|, \\ &\text{with} \\ &D_{\mathbf{w}}(s) = \sigma \left(\text{logit}(O(s)) + RD(s)\right) \text{ and} \\ &RD(s) = w\text{ReLU}(w_{SR'}s + \text{bias}_{R'}).
\end{align*}
Here, $R'$ is the representational disparity node; $O(s) = \text{Pr}_\text{obs}(H=1\:|\: S=s)$, $RD(s)$ measures the representational disparity; $\mathbf{w}$ is comprised of $(w, w_{SR'}, \text{bias}_{R'})$; and $\sigma$ is the sigmoid function.\\[1ex]
We prove that initializing $\mathbf{w}$ to non-zero values $\{w > 0, w_{SR'} > 0, \text{bias}_{R'} \ge -w_{SR'}\}~\text{for}~\delta < 0$, or $\{w < 0, w_{SR'} > 0, \text{bias}_{R'} \ge -w_{SR'}\}~\text{for}~\delta > 0$,
and using gradient descent results in the global optimum loss $L_{\text{min}}$ attained at $\mathbf{w}_{\text{min}}$ given by,
\begin{align*}
L_{\text{min}} &= 2\sqrt{|\delta|} \\
\mathbf{w}_{\text{min}} &= \{w=-\text{sign}(\delta)\sqrt{|\delta|}, w_{SR'}=\sqrt{|\delta|}, \text{bias}_{R'} = 0\}
\end{align*}
\end{theorem}
The weights to which the network converges are interpretable. For instance, let the observed human decision allocating housing vouchers discriminate against the protected subgroup $S=1$ with $\delta < 0$. Then, the representational disparity $RD(1)-RD(0) = ww_{SR'} = -\delta$ implies that the learned weights compensate for the existing discrimination against the subgroup $S=1$. 

The proof of Theorem 4.1 is based on finding weights that make $A_\mathbf{w}=0$, and then minimizing $B_\mathbf{w}$ among the weights that make $A_\mathbf{w}=0$ since $a \gg b$. We show $A_\mathbf{w}=0$ if and only if $w[\text{ReLU}(\text{bias}_{R'}) - \text{ReLU}(w_{SR'} + \text{bias}_{R'})] = \delta$. We divide the search space comprising $\{w, w_{SR'}, \text{bias}_{R'}\}$ into regions based on the signs of these weights. 
%With non-sensitive attributes $\mathbf{X}$, as in Theorem 4.2, such a division is not possible as ReLU cannot be simplified. 
We show that restricting the search space to these feasible regions makes the optimization problem convex with a strictly convex and continuous objective, which guarantees unique local minima~\cite{boyd2004convex}. We can thus initialize weights within regions that result in the global minimum loss. 

\textbf{Theorem 4.2} relaxes assumptions (A4) and (A5), allowing non-sensitive attributes $\mathbf{X}$ and multiple representational disparity nodes.  Under assumptions (A1)-(A3) and (A6), the globally optimal weights fully mitigate the outcome disparity and remain interpretable, with only a single node used to mitigate disparity.  \textbf{Theorem 4.3} instead relaxes assumption (A6), allowing weight of disparity loss $a$ and interpretability loss $b$ to be similar in magnitude.  In this case, there are several possible solutions for the globally optimal weights (shifting one group's probability $\text{Pr}(H=1\:|\: S)$ toward the other or pushing both probabilities to an extreme), and the observed disparity may be partially rather than totally mitigated.
We note that, unlike Theorem 4.1, Theorems 4.2 and 4.3 do not prove the network's convergence to the global minimum loss.  See Appendix~\ref{proofs} for proofs.

\textbf{Theorem 4.2}. Under the same preconditions as Theorem 4.1, including assumptions (A1)-(A3) and (A6), but with the added complexity of having non-sensitive attributes $\mathbf{X}$ and training with multiple ($k > 1$) disparity nodes, Theorem 4.2 proves that optimal weights are interpretable with only a single node being used to mitigate disparity. Here we again assume that there is $\delta$-unfairness towards $S=1$ in the observed decision $H$, $\delta = \text{logit}(H=1\:|\: \mathbf{X=x}, S=1)-\mbox{logit}(H=1\:|\:\mathbf{X=x},S=0)$ for all $\mathbf{x}$, $\delta \ne 0$. In this case, the global minimum loss $L_{\text{min}}$ attained at $\mathbf{w}_{\text{min}}$ is,
\begin{align*}
&L_{\text{min}} = 2\sqrt{|\delta|} \\
&\mathbf{w}_{\text{min}} = \{\exists i \in \{1, ... , k\} ~\text{s.t.}, w_i=-\text{sign}(\delta)\sqrt{|\delta|}, \mathbf{w}_{\mathbf{X}R'_i}=0, w_{SR'_i}=\sqrt{|\delta|}, \text{bias}_{R'_i}=0, \nonumber \\ &~~~~~~~~~~~~~~~\forall j \neq i, j \in \{1, ... , k\}, w_j=0, \mathbf{w}_{\mathbf{X}R'_j}=0, w_{SR'_j}=0, \text{bias}_{R'_j}=0\}
\end{align*}
The weights to which the network converges remain interpretable. For instance, let the observed human decision allocating housing vouchers discriminate against the protected subgroup $S=1$ with $\delta < 0$. Then, the representational disparity $RD(\mathbf{x},1) - RD(\mathbf{x},0) = ww_{SR'} = -\delta$ for all $\mathbf{x}$ implies that the learned weights compensate for the existing discrimination against the subgroup $S=1$. 

The proof is again based on finding weights that make $A_\mathbf{w}=0$, and then finding the minimum $B_\mathbf{w}$ among weights that make $A_\mathbf{w}=0$ since $a \gg b$. We consider two cases that make $A_\mathbf{w}=0$, one with 
$D_\mathbf{w}(\mathbf{x},1)=D_\mathbf{w}(\mathbf{x},0)$ 
%$\Delta D_\mathbf{w}(\mathbf{x})=0$ 
for all $\mathbf{x}$, and another with $\exists \mathbf{x}~\text{such that}~D_\mathbf{w}(\mathbf{x},1) \ne D_\mathbf{w}(\mathbf{x},0)$, and show that the former results in the minimum loss, implying that the desired decision-maker does not take $\mathbf{X}$ into account when making decisions. A trivial inequality is used to show that only one node is used to mitigate disparity.

\textbf{Theorem 4.3}. Under the same preconditions as Theorem 4.1, including assumptions (A1)-(A5), but with the added complexity that disparity loss is comparable to interpretability loss ($a \approx b$), Theorem 4.3 proves that the optimal losses and weights remain interpretable. The global minimum loss $L_{\text{min}}$ attained at $\mathbf{w}_{\text{min}}$ is,
\begin{align*}
L_{\text{min}} &= \left\{\begin{array}{lr}
        \text{min}\{\underbrace{\underset{B_\mathbf{w}}{\text{min}}~(1-a)B_\mathbf{w} + a|\alpha|SD\left(\frac{B_\mathbf{w}^2}{4}\right)}_{L1}, ~\underbrace{\underset{B_\mathbf{w}}{\text{min}}~(1-a)B_\mathbf{w} + a|\alpha|EI\left(\frac{B_\mathbf{w}^2}{4}\right)\}}_{L2}, &~\delta > 0\\
        \text{min}\{\underbrace{\underset{B_\mathbf{w}}{\text{min}}~(1-a)B_\mathbf{w} + a|\alpha|SI\left(\frac{B_\mathbf{w}^2}{4}\right)}_{L3}, ~\underbrace{\underset{B_\mathbf{w}}{\text{min}}~(1-a)B_\mathbf{w} + a|\alpha|ED\left(\frac{B_\mathbf{w}^2}{4}\right)\}}_{L4} &~\delta < 0
        \end{array}\right. \\
\mathbf{w}_{\text{min}} &= \left\{\begin{array}{lr}w = -\frac{B_\text{opti}}{2}, w_{SR'} = \frac{B_\text{opti}}{2}, \text{bias}_{R'}=0, \text{when}~\text{loss $L1$ is chosen} \\ w = \frac{B_\text{opti}}{2}, w_{SR'} = 0, \text{bias}_{R'}= \frac{B_\text{opti}}{2}, \text{when}~\text{loss $L2$ is chosen}\\ w = \frac{B_\text{opti}}{2}, w_{SR'} = \frac{B_\text{opti}}{2}, \text{bias}_{R'}=0, \text{when}~\text{loss $L3$ is chosen}\\ w = - \frac{B_\text{opti}}{2}, w_{SR'} = 0, \text{bias}_{R'} = \frac{B_\text{opti}}{2}, \text{when}~\text{loss $L4$ is chosen} \end{array}\right.
\end{align*}
where $B_\text{opti}$ is the optimal $B_\mathbf{w}$; $SD$ is a decrease in logit of the sigmoid with the larger logit; $EI$ is an equal increase in logit; $SI$ is an increase in logit of the sigmoid with the smaller logit; and $ED$ is an equal decrease in logit. See Appendix~\ref{proofs} for equations of $SD$, $EI$, $SI$, and $ED$.

We can find the minimum loss by numerical methods. The losses learned are interpretable. For instance, let the observed human decision of allocating housing vouchers favor the protected subgroup $S=1$ with $\delta > 0$. Then, the disparity loss can be mitigated either by decreasing the logit of the sigmoid with the larger logit, as in $SD(x)$, or by increasing the logit of both the sigmoids equally, in other words, pushing both probabilities toward 1 to decrease the difference between them, as in $EI(x)$. Similar analysis can be made for $\delta < 0$. Further, the weights learned are interpretable. When loss $L1$ ($\delta > 0$) or loss $L3$ ($\delta < 0$) is chosen, the weights learned to compensate for the existing favoritism by making $ww_{SR'}<0$ when $\delta > 0$, or compensate for the existing discrimination by making $ww_{SR'}>0$ when $\delta < 0$. Similar analysis can be made for losses $L2$ ($\delta > 0$) and $L4$ ($\delta < 0$). 

The proof differs from the above two theorems as $a$, the hyper-parameter of $A_\mathbf{w}$, is comparable to $1-a$, the hyper-parameter of $B_\mathbf{w}$. We re-write the optimization problem $aA_\mathbf{w} + (1-a)B_\mathbf{w}$ as a two-step procedure: first to find the minimum disparity $A_\mathbf{w}$ as a function of $B_w$, and then to minimize the total loss with respect to $B_\mathbf{w}$. The challenges are two-fold: to write $A_\mathbf{w}$ in terms of $B_\mathbf{w}$ and to derive interpretable weights. Here, we prove the theorem for a simplified setting comprising of only the sensitive attribute $S$, as writing $A_\mathbf{w}$ in terms of $B_\mathbf{w}$ is non-trivial when non-sensitive attributes $\mathbf{X}$ and its weights $w_{\mathbf{X}R'}$ are involved.  

\section{Experiments}
\label{exp_sec}

%We validate the optimization objectives using synthetic datasets. Since convergence to the global optimum is guaranteed in Theorem 4.1, we only validate Theorem 4.2 and 4.3 using synthetic datasets to convince that the global optimum is achieved in practice (see Appendix \ref{expdetails}).

\textbf{Datasets} \\
We present evaluation results on three real-world datasets: German Credit, Adult income, and Heritage Health. The German Credit dataset classifies bank holders into a Good or Bad credit class. We use $Age$ as the sensitive attribute, following \cite{zemel2013learning, kamiran2009classifying}. The Adult income dataset classifies whether or not each individual's income is above \$50,000. We use $Gender$ as the sensitive attribute, following \cite{zemel2013learning, kohavi1996scaling, kamishima2012fairness}. The Heritage Health dataset classifies whether each patient spends any days in the hospital that year. We use $Age$ as the sensitive attribute, following \cite{zemel2013learning}. In each dataset, all attributes are binarized by a one-hot encoding of categorical attributes and quantization of numerical attributes. Data is split 70\% for training and 30\% for testing. We report results on the test data, and use 10-fold cross-validation within the training data for model selection. See Appendix~\ref{expdetails} for details.

\textbf{Comparator methods}\\
We compare our approach (LRD) with a competing approach, Learning Fair Representations (LFR)~\cite{zemel2013learning}. We reproduce LFR by modifying Prof. Zubin Jelveh's implementation~\cite{lfrimplem}. We validate our LFR implementation by matching the accuracy ($yAcc$) and outcome disparity ($yDisc$) values reported in Table 1 of~\cite{zemel2013learning} for the Adult and German Credit datasets. %LRD training is described below. We report results on the test data.

\textbf{Experiments}\\
As noted above, we are interested in mitigating disparities in a downstream outcome $Y$ affected by the human decision $H$. We note that the distribution of $Y$ given $H$ is assumed to be fixed and cannot be changed by the methods; they can only modify the human decision $H$ to compensate for existing biases in $H$ and in $Y$ given $H$.  Since the three real-world datasets described above do not have a downstream outcome that is separate from the class variable to be predicted, we perform five different semi-synthetic experiments which use the class variable as the human decision $H$, and generate a new outcome variable $Y$ which is dependent on $H$. (Note that, if $Y$ was independent of $H$, the outcome disparity could not be reduced by modifying $H$.)
To do so, we decompose the total outcome disparity, $|\text{Pr}(Y=1\:|\:S=1)-\text{Pr}(Y=1\:|\:S=0)|$, as $|ac+b|$, where $a = 
\text{Pr}(Y=1\:|\:S=s,H=1)-\text{Pr}(Y=1\:|\:S=s,H=0)$ for all $s$; 
$b = \text{Pr}(Y=1\:|\:S=1,H=h)-\text{Pr}(Y=1\:|\:S=0,H=h)$ for all $h$; and $c = \text{Pr}(H=1\:|\:S=1)-\text{Pr}(H=1\:|\:S=0)$.  We fix $a$ to a constant for simplicity, while $c$ is dataset-dependent.  We then formulate five cases using different values of $b$ (see Appendix~\ref{expdetails} for full details):\\[1ex]
\textbf{Case I:} The disparity (between $S=1$ and $S=0$) in the outcome process $Y \:|\: H$ adds to the existing disparity in $Y$ resulting from disparity in $H$, achieved by setting $b=ac$.\\
\textbf{Case II:} The disparity in the outcome process counteracts the existing disparity in $Y$ resulting from disparity in $H$, but does not fully eliminate that disparity, achieved by setting $b=-0.5ac$.\\ \textbf{Case III:} The disparity in 
the outcome process counteracts and fully eliminates the existing disparity in $Y$ resulting from disparity in $H$, achieved by setting $b=-ac$.\\
\textbf{Case IV:} The disparity in the outcome process overwhelms and reverses the direction of the existing disparity in $Y$ resulting from disparity in $H$, achieved by setting $b=-1.5ac$.\\
\textbf{Case V:} There is no disparity in the outcome process ($b=0$).\\[1ex] 
By setting $a=1$ in Case V, we consider the special case where $Y=H$.
For the remaining Cases I-IV, we set 
$a=0.6$.

\textbf{Model Selection}\\
We train the architecture shown in Figure~\ref{neural_network} using 5-fold cross-validation on the training data \{($\mathbf{X}$, $S$, $H$)\} to select the number of nodes used to model the observed decision-maker ($m'$). The number of nodes selected remains the same for Cases I to V. Total loss vs. $m'$ plots are shown in Appendix~\ref{expdetails}. Based on the results, $m'=1$ for the German Credit dataset, $m'=4$ for the Adult dataset, and $m'=11$ for the Health dataset. For maximal interpretability, we use a single additional node to capture the representational disparity ($m=m'+1$). 

\textbf{Training}\\
We train 100 fits only on Objectives $C$ and $D$, which corresponds to learning the observed and outcome processes, and a fit with minimum total loss is selected to freeze the weights of the observed and outcome decision processes. We then train using Adam optimizer \cite{kingma2014adam} to minimize Eq.~\ref{totalloss1}. We train 100 fits by setting $a=0.99$, $b=0.01$, $c=1000$ and $d=1000$, and select the fit with minimum training loss. We chose $a \gg b$ for maximal reduction of the outcome disparity. %See Appendix \ref{expdetails} for details. (Commented out after main paper submission DBN)

\textbf{Results}\\
For outcome disparity and accuracy, results are averaged across 10 train-test splits and reported in Table~\ref{keyresults}. Outcome disparity is formulated in Eq. \ref{initform} and (equivalently) in Eq.~14 of~\cite{zemel2013learning}. LRD achieves substantially reduced disparity in Cases I to IV, and similar disparity in Case V, compared to LFR.  We observe that LFR is only able to remove the impact of $S$ on $H$ (resulting in $c\approx 0$ and final disparity $\approx |b|$); while LRD also accounts for disparities between $S=1$ and $S=0$ in the outcome process $\text{Pr}(Y=1 \:|\: S, H)$. We note two cases where a substantial amount of disparity remains after correction: first, for the German Credit dataset, LRD removes essentially all of the disparity from the training data, and the remaining disparity is due to small dataset size (1000 records) and differences between training and test partitions.  Second, for Case I in the Adult dataset, it is not possible to counteract all of the disparity in the outcome process by modifying $H$ alone.  Finally, we note that, while numerous recent variants of LFR have been proposed, these methods would all perform similarly to LFR (and thus, underperform LRD) since they aim to make the human decision $H$ independent of $S$, and do not account for the downstream disparity in the outcome $Y$ given $H$. 

Accuracy using fair predictions is formulated in Eq. 13 of~\cite{zemel2013learning}. This can be interpreted as the proportion of test samples in which the desired human decision matches the observed human decision. LRD consistently achieves higher accuracy for $H$ than LFR across all five experiments.  We believe that this improvement may result from a better model of the observed decision-maker (choosing the number of representation nodes by cross-validation) as well as the increased consistency in how the desired and observed decisions differ (as discussed below).

%, because, for every dataset, LRD consciously selects the number of representational nodes used by the observed decisioner using 5-cross validation. In addition, deviation from the observed behavior is varied smoothly using additional nodes to mitigate the disparity. However, LFR sets the number of prototypes (analogous to the representational nodes that capture the desired decisions) to an arbitrary value of $k=10$ for every dataset \footref{fn1}.

\begin{table} [t]
\begin{center}
\begin{tabular}{@{}p{2.5cm} p{1.2cm}p{1.2cm}p{1.2cm}p{1.2cm}p{1.2cm}@{}}
\hline
& \multicolumn{3}{c}{\textbf{Outcome Disparity}} & \multicolumn{2}{c}{\textbf{Accuracy}} \\ & \textbf{LRD} & \textbf{|b|} & \textbf{LFR} & \textbf{LRD} & \textbf{LFR}  \\
\hline
\textbf{Case I} ~~~German &0.0187 &0.0565 &0.0614	&0.6092 &0.5839 \\ {~~~~~~~~~~~~~~~Adult} &0.0698 &0.1141 &0.1199	&0.5524 &0.5395 \\ {~~~~~~~~~~~~~~~Health} &0.0012 &0.0530 &0.0544	&0.5640 &0.5303 \\ \textbf{Case II} ~~German &0.0165 &0.0291 &0.0247 &0.6279 &0.5872 \\ {~~~~~~~~~~~~~~~Adult} &0.0053 &0.0573 &0.0518	&0.5638 &0.5266 \\ {~~~~~~~~~~~~~~~Health} &0.0024 &0.0266 &0.0253 &0.5583 &0.5184 \\ \textbf{Case III} German &0.0171 &0.0603 &0.0554 &0.6173 &0.5753 \\ {~~~~~~~~~~~~~~~Adult} &0.0035 &0.1172 &0.1114	&0.5882 &0.5439 \\ {~~~~~~~~~~~~~~~Health} &0.0013 &0.0533 &0.0520	&0.5610 &0.5229 \\ \textbf{Case IV} German &0.0217 &0.0835 &0.0786	&0.5991 &0.5740\\ {~~~~~~~~~~~~~~~Adult} &0.0050 &0.1726 &0.1669	&0.6040 &0.5705 \\ {~~~~~~~~~~~~~~~Health} &0.0031 &0.0794 &0.0781 &0.5656 &0.5285 \\ \textbf{Case V} ~~German &0.0244 &0 &0.0148 &0.6574 &0.5938 \\ {~~~~~~~~~~~~~~~Adult} & 0.0053 &0 &0.0124 & 0.7670 & 0.6549 \\ {~~~~~~~~~~~~~~~Health} &0.0031 &0 &0.0022 &0.8464 &0.6899
\end{tabular} 
\caption{LRD and LFR Results Comparison}
\label{keyresults}
\end{center}
\end{table}

We make two additional points that LRD improves on LFR in both consistency and interpretability. For a given experiment and data split, LRD consistently shifts one class's probabilities of $\text{Pr}(Y=1\:|\:S=s)$ toward the other, with little change to the other class's probabilities. In contrast, LFR creates wide variation in individual probabilities: many observations in each class have substantial increases and substantial decreases in probability. We note that different train/test splits can result in either the lower-probability class having its probabilities shifted upward, or the higher-probability class having its probabilities shifted downward, by LRD. Similar consistency results were seen for all cases, but we note that in Case III, LRD made no corrections to $H$ (all weights for the representational disparity node were very close to 0).  In this case, no corrections were necessary as the disparities in $H$ and $Y\:|\:H$ cancel out. 

As for interpretability, for Cases I, II, IV, and V, across all splits, LRD placed greatest weight on $age$, $relationship\_Husband$, and $ageGrEq65$ for the German Credit, Adult, and Health datasets, respectively. For Cases I, II, IV and V, the correction (sign of the product of incoming and outgoing weights to the representational disparity node) indicates that it reduces the disparity in $Y$. For Case III, for Adult and Health datasets all corrections are near 0 as there is no disparity in $Y$; for German Credit there was a small but non-zero correction resulting from small data size and differences between train and test partitions. While $age$ and $ageGrEq65$ are sensitive attributes, LRD's use of $relationship\_Husband$ for the Adult dataset is notable, as the gender disparity in $income$ was heavily impacted by an individual's marital status. For 18,986 men with $relationship\_Husband$ = 1, 45\% had $income$ = 1; for 12,255 men with $relationship\_Husband$ = 0, only 9\% had $income$ = 1, similar to the proportion for women. 

Finally, we explain how disparity is mitigated. Consider a split of the Adult data in Case II with $\text{Pr}_\text{obs}(Y=1\:|\:S=1)=0.3287$ and $\text{Pr}_\text{obs}(Y=1\:|\:S=0)=0.5338$. Here, the outcome $Y=1$ (e.g., awarding of a loan) and the human decision $H=1$ (``income greater than or equal to 50K'') both favor males ($S=0$). The neural network learns to use $relationship\_Husband$ with $w_{AR'}w=-0.7049$ contribution to decrease the logits of $\text{Pr}(Y=1\:|\:S=0)$ and reduce the gender disparity in outcomes.

\section{Conclusion}
We propose a novel algorithm to model the disparities between the observed and fair decision-makers. We validate each of the training objectives and prove that the weights learned are interpretable. Using real-world datasets, we investigate the disparities and demonstrate the effectiveness of our approach by comparing with a foundational work. 
Our future work will examine various extensions and generalizations of the proposed LRD approach, including (i) multiple sensitive attributes and intersectional subgroups; (ii) approaches that do not use the sensitive attribute for legal compliance; and (iii) hybrid models that use deeper, more complex networks to model the observed human decision while maintaining easily interpretable and actionable representational disparities.

%\printbibliography
\bibliographystyle{plain}
\bibliography{ref}

\begin{thebibliography}{10}

\bibitem{angwin2016machine}
Julia Angwin, Jeff Larson, Surya Mattu, and Lauren Kirchner.
\newblock Machine bias. propublica, may 23, 2016, 2016.

\bibitem{barocas2017fairness}
Solon Barocas, Moritz Hardt, and Arvind Narayanan.
\newblock Fairness in machine learning.
\newblock {\em Nips tutorial}, 1:2017, 2017.

\bibitem{black2023toward}
Emily Black, Rakshit Naidu, Rayid Ghani, Kit Rodolfa, Daniel Ho, and Hoda Heidari.
\newblock Toward operationalizing pipeline-aware ml fairness: A research agenda for developing practical guidelines and tools.
\newblock In {\em Proceedings of the 3rd ACM Conference on Equity and Access in Algorithms, Mechanisms, and Optimization}, pages 1--11, 2023.

\bibitem{boyd2004convex}
Stephen~P Boyd and Lieven Vandenberghe.
\newblock {\em Convex optimization}.
\newblock Cambridge university press, 2004.

\bibitem{marketmarkers}
Phil Brierley, David Vogel, and Randy Axelrod.
\newblock Heritage provider network health prize round 1 milestone prize how we did it – team "market makers".
\newblock {\em foreverdata.org/1015/content/milestone1-2.pdf}, 2011.

\bibitem{de2018clinically}
Jeffrey De~Fauw, Joseph~R Ledsam, Bernardino Romera-Paredes, Stanislav Nikolov, Nenad Tomasev, et~al.
\newblock Clinically applicable deep learning for diagnosis and referral in retinal disease.
\newblock {\em Nature Medicine}, 24(9):1342--1350, 2018.

\bibitem{eichenbaum2014can}
Howard Eichenbaum and Neal~J Cohen.
\newblock Can we reconcile the declarative memory and spatial navigation views on hippocampal function?
\newblock {\em Neuron}, 83(4):764--770, 2014.

\bibitem{Gelman01092007}
Andrew Gelman, Jeffrey Fagan, and Alex Kiss.
\newblock An analysis of the new york city police department's “stop-and-frisk” policy in the context of claims of racial bias.
\newblock {\em Journal of the American Statistical Association}, 102(479):813--823, 2007.

\bibitem{goodfellow2016deep}
Ian Goodfellow, Yoshua Bengio, and Aaron Courville.
\newblock {\em Deep learning}.
\newblock MIT press, 2016.

\bibitem{green2020algorithm}
Ben Green and Yiling Chen.
\newblock Algorithm-in-the-loop decision making.
\newblock In {\em Proceedings of the AAAI Conference on Artificial Intelligence}, pages 13663--13664, 2020.

\bibitem{lfrimplem}
Zubin Jelveh.
\newblock Learning fair representations implementation.
\newblock {\em github.com/zjelveh/learning-fair-representations}, 2015.

\bibitem{kamiran2009classifying}
Faisal Kamiran and Toon Calders.
\newblock Classifying without discriminating.
\newblock In {\em 2009 2nd international conference on computer, control and communication}, pages 1--6. IEEE, 2009.

\bibitem{kamishima2012fairness}
Toshihiro Kamishima, Shotaro Akaho, Hideki Asoh, and Jun Sakuma.
\newblock Fairness-aware classifier with prejudice remover regularizer.
\newblock In {\em Machine Learning and Knowledge Discovery in Databases: European Conference, ECML PKDD 2012, Bristol, UK, September 24-28, 2012. Proceedings, Part II 23}, pages 35--50. Springer, 2012.

\bibitem{kingma2014adam}
Diederik~P Kingma and Jimmy Ba.
\newblock Adam: A method for stochastic optimization.
\newblock {\em arXiv preprint arXiv:1412.6980}, 2014.

\bibitem{kohavi1996scaling}
Ron Kohavi et~al.
\newblock Scaling up the accuracy of naive-bayes classifiers: A decision-tree hybrid.
\newblock In {\em Kdd}, volume~96, pages 202--207, 1996.

\bibitem{koller2009probabilistic}
Daphne Koller and Nir Friedman.
\newblock {\em Probabilistic graphical models: principles and techniques}.
\newblock MIT press, 2009.

\bibitem{luo2023learning}
Yuhong Luo, Austin Hoag, and Philip~S Thomas.
\newblock Learning fair representations with high-confidence guarantees.
\newblock {\em arXiv preprint arXiv:2310.15358}, 2023.

\bibitem{madras2018learning}
David Madras, Elliot Creager, Toniann Pitassi, and Richard Zemel.
\newblock Learning adversarially fair and transferable representations.
\newblock In {\em International Conference on Machine Learning}, pages 3384--3393. PMLR, 2018.

\bibitem{mehrabi2021survey}
Ninareh Mehrabi, Fred Morstatter, Nripsuta Saxena, Kristina Lerman, and Aram Galstyan.
\newblock A survey on bias and fairness in machine learning.
\newblock {\em ACM Computing Surveys (CSUR)}, 54(6):1--35, 2021.

\bibitem{ravishankar2023provable}
Pavan Ravishankar, Qingyu Mo, Edward McFowland~III, and Daniel~B Neill.
\newblock Provable detection of propagating sampling bias in prediction models.
\newblock {\em Association for the Advancement of Artificial Intelligence, 2023}, 2023.

\bibitem{smith2021mental}
Stephanie~M Smith and Ian Krajbich.
\newblock Mental representations distinguish value-based decisions from perceptual decisions.
\newblock {\em Psychonomic Bulletin \& Review}, 28:1413--1422, 2021.

\bibitem{squire1991medial}
Larry~R Squire and Stuart Zola-Morgan.
\newblock The medial temporal lobe memory system.
\newblock {\em Science}, 253(5026):1380--1386, 1991.

\bibitem{suresh2021framework}
Harini Suresh and John Guttag.
\newblock A framework for understanding sources of harm throughout the machine learning life cycle.
\newblock In {\em Proceedings of the 1st ACM Conference on Equity and Access in Algorithms, Mechanisms, and Optimization}, pages 1--9, 2021.

\bibitem{suzuki2009memory}
Wendy~A Suzuki and Mark~G Baxter.
\newblock Memory, perception, and the medial temporal lobe: a synthesis of opinions.
\newblock {\em Neuron}, 61(5):678--679, 2009.

\bibitem{thaler2009nudge}
Richard~H Thaler and Cass~R Sunstein.
\newblock {\em Nudge: Improving decisions about health, wealth, and happiness}.
\newblock Penguin, 2009.

\bibitem{zemel2013learning}
Rich Zemel, Yu~Wu, Kevin Swersky, Toni Pitassi, and Cynthia Dwork.
\newblock Learning fair representations.
\newblock In {\em International conference on machine learning}, pages 325--333. PMLR, 2013.

\bibitem{zhao2019conditional}
Han Zhao, Amanda Coston, Tameem Adel, and Geoffrey~J Gordon.
\newblock Conditional learning of fair representations.
\newblock {\em arXiv preprint arXiv:1910.07162}, 2019.

\bibitem{zhao2022inherent}
Han Zhao and Geoffrey~J Gordon.
\newblock Inherent tradeoffs in learning fair representations.
\newblock {\em Journal of Machine Learning Research}, 23(57):1--26, 2022.

\end{thebibliography}

\clearpage
\appendix
\section{Proofs}
\label{proofs}
\textbf{Theorem 4.1} Assume the data generating process and neural network architecture in Figures~\ref{datagengraph}-\ref{neural_network} and assumptions (A1)-(A6) above. Here the decision $H$ depends only on $S$, and the outcome $Y$ depends only on $H$. Let $\alpha = \text{Pr}(Y=1\:|\: H=1)-\mbox{Pr}(Y=1\:|\:H=0)$, $\alpha \ne 0$.  Moreover, assume that there is $\delta$-unfairness towards $S=1$ in the observed decision $H$, $\delta = \text{logit}(H=1\:|\: S=1)-\text{logit}(H=1\:|\:S=0)$, $\delta \ne 0$. Suppose the desired decision $D_{\mathbf{w}}(s) = \text{Pr}_\text{des}(H=1 \:|\: S=s)$, parameterized by weight vector $\mathbf{w} = (w, w_{SR'},\text{bias}_{R'})$, is learned using training data $T$ by minimizing the total loss $L_{\mathbf{w}}$, where
\begin{align*}
&L_{\mathbf{w}} = aA_{\mathbf{w}} + bB_{\mathbf{w}}, \\
&A_{\mathbf{w}} = |\alpha||D_{\mathbf{w}}(1)-D_{\mathbf{w}}(0)|, \text{ and} \\  &B_{\mathbf{w}} = |w| + |w_{SR'}| + |\text{bias}_{R'}|, \\ &\text{with} \\ &D_{\mathbf{w}}(s) = \sigma \left(\text{logit}(O(s)) + RD(s)\right) \text{ and} \\ &RD(s) = w\text{ReLU}(w_{SR'}s + \text{bias}_{R'}).
\end{align*}
Here, $R'$ is the representational disparity node; $O(s) = \text{Pr}_\text{obs}(H=1\:|\: S=s)$, $RD(s)$ measures the representational disparity; $\mathbf{w}$ is comprised of $(w, w_{SR'}, \text{bias}_{R'})$; and $\sigma$ is the sigmoid function.\\[1ex]
We prove that initializing $\mathbf{w}$ to non-zero values $\{w > 0, w_{SR'} > 0, \text{bias}_{R'} \ge -w_{SR'}\}~\text{for}~\delta < 0$, or $\{w < 0, w_{SR'} > 0, \text{bias}_{R'} \ge -w_{SR'}\}~\text{for}~\delta > 0$,
and using gradient descent results in the global optimum loss $L_{\text{min}}$ attained at $\mathbf{w}_{\text{min}}$ given by,
\begin{align*}
L_{\text{min}} &= 2\sqrt{|\delta|} \\
\mathbf{w}_{\text{min}} &= \{w=-\text{sign}(\delta)\sqrt{|\delta|}, w_{SR'}=\sqrt{|\delta|}, \text{bias}_{R'} = 0\}
\end{align*}

\textit{Proof:} 
\noindent Since $a \gg b$, we minimize $B_\mathbf{w}$ under the constraint that $A_\mathbf{w} = 0$.  Since $\alpha \ne 0$, this implies $D_{\mathbf{w}}(1) = D_{\mathbf{w}}(0)$, and therefore
$\sigma(\text{logit}(O(1)) + w\text{ReLU}(w_{SR'} + \text{bias}_{R'})) =\sigma(\text{logit}(O(0)) + w\text{ReLU}(\text{bias}_{R'}))$.  Under the $\delta$-unfairness assumption above, this implies:
\begin{align}
w[\text{ReLU}(\text{bias}_{R'}) - \text{ReLU}(w_{SR'} + \text{bias}_{R'})] = \delta. \label{opticons}
\end{align}
\noindent Now, we find the $\mathbf{w}=\{w,w_{SR'}, \text{bias}_{R'}\}$ that minimizes $B_{\mathbf{w}}$:
\begin{align}
    &\underset{w, w_{SR'}, \text{bias}_{R'}} {\text{min}} ~|w| + |w_{SR'}| + |\text{bias}_{R'}| \nonumber\\
    &\text{s.t.,} ~w[\text{ReLU}(\text{bias}_{R'}) - \text{ReLU}(w_{SR'} + \text{bias}_{R'})] = \delta. \label{reluacti1}
\end{align}

\begin{figure}[t]
    \centering
    \includegraphics[width=0.5\linewidth]{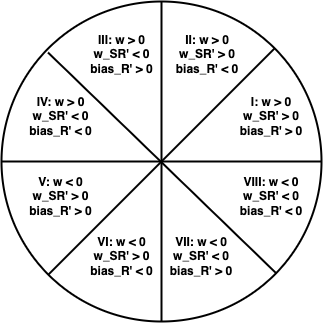}
    \caption{Regions divided based on the signs of $w$, $w_{SR'}$, and $\text{bias}_{R'}$.}
    \label{fig:quad}
\end{figure}

We divide the search space comprising $\{w, w_{SR'}, \text{bias}_{R'}\}$ into regions based on the signs of these weights as shown in Figure~\ref{fig:quad}. With non-sensitive attributes $\mathbf{X}$, as in Theorem 4.2 below, such a division is not possible as the ReLU in Equation~\ref{reluacti1} cannot be simplified. We show that restricting the search space to these feasible regions makes the optimization problem convex with a strictly convex and continuous objective, resulting in unique local minima. Suppose $w > 0$, $w_{SR'} > 0$, $\text{bias}_{R'} > 0$ and $\delta < 0$. Then,
\begin{align*}
    &~~~~~\underset{w, w_{SR'}, \text{bias}_{R'}} {\text{min}} ~w + w_{SR'} + \text{bias}_{R'} \\
    &~~~~~\text{s.t.,} ~-ww_{SR'} = \delta, w > 0, w_{SR'} > 0, \text{bias}_{R'} > 0 \\
    &\equiv \underset{w > 0, \text{bias}_{R'} > 0} {\text{min}} ~w + \frac{|\delta|}{w} + \text{bias}_{R'} \\
    &\equiv \underset{w > 0} {\text{min}} ~w + \frac{|\delta|}{w} + \underset{\text{bias}_{R'} > 0} {\text{min}} \text{bias}_{R'}
\end{align*}
whose only local minima loss in $\{w > 0$, $w_{SR'} > 0$, $\text{bias}_{R'} > 0\}$ region is $2\sqrt{|\delta|}$ with $\{w=\sqrt{|\delta|}, w_{SR'}=\sqrt{|\delta|}, \text{bias}_R' \rightarrow 0\}$. This local optimum can be reached by initialization $\{w, w_{SR'}, \text{bias}_R'\}$ to any point in the space $\{w > 0$, $w_{SR'} > 0$, $\text{bias}_{R'} > 0$\} as $w + \frac{|\delta|}{w}$ is strictly convex and continuous. $\underset{w > 0} {\text{min}} ~w + \frac{|\delta|}{w}$ and $\underset{\text{bias}_{R'} > 0} {\text{min}} \text{bias}_{R'}$ are separately calculated as $\text{bias}_{R'}$ is not dependent on $w$.

Similarly, one can derive the minimum loss for other regions. When $\delta < 0$,  for $\{w > 0, w_{SR'} > 0, \text{bias}_{R'} < 0, w_{SR'} + \text{bias}_{R'} \ge 0\}$, the only local minimum loss in the region is $2\sqrt{|\delta|}$ with $\{w=\sqrt{|\delta|}, w_{SR'}=\sqrt{|\delta|}, \text{bias}_{R'} \rightarrow 0\}$. When $\delta > 0$, for $\{w < 0, w_{SR'} > 0, \text{bias}_{R'} > 0\}$, the only local minimum loss in the region is $2\sqrt{|\delta|}$ with $\{w \rightarrow -\sqrt{|\delta|}, w_{SR'} = \sqrt{|\delta|}, \text{bias}_{R'} \rightarrow 0\}$; for $\{w < 0, w_{SR'} > 0, \text{bias}_{R'} < 0, w_{SR'} + \text{bias}_{R'} \geq 0\}$, the only local minimum loss in the region is $2\sqrt{|\delta|}$ with $\{w \rightarrow -\sqrt{|\delta|}, w_{SR'} = \sqrt{|\delta|}, \text{bias}_{R'} \rightarrow 0\}$. 

For other feasible regions, the only local minimum loss is $2\sqrt{2|\delta|}$ with $\{|w|=\sqrt{2|\delta|}, |w_{SR'}|=\sqrt{\frac{|\delta|}{2}}, |\text{bias}_{R'}| = \sqrt{\frac{|\delta|}{2}}\}$ with the signs dictated by their search regions.

There is no feasible solution when $\{w = *, w_{SR'} = *, \text{bias}_{R'} < 0, w_{SR'} + \text{bias}_{R'} < 0\}$ for any $\delta$; $\{w > 0, w_{SR'} > 0, \text{bias}_{R'} > 0\}$ and $\{w < 0, w_{SR'} < 0, \text{bias}_{R'} > 0, w_{SR'} + \text{bias}_{R'} < 0\}$ for $\delta > 0$; $\{w > 0, w_{SR'} < 0, \text{bias}_{R'} > 0\}$ and $\{w < 0, w_{SR'} > 0, \text{bias}_{R'} > 0\}$ and $\{w < 0, w_{SR'} > 0, \text{bias}_{R'} < 0, w_{SR'} + \text{bias}_{R'} \geq 0\}$ for $\delta < 0$ as the optimization constraint (Eq. \ref{opticons}) becomes inconsistent ($*$ means any value taken). 

Hence, initializing $\{w, w_{SR'}, \text{bias}_{R'}\}$ to non-zero values in $\{w > 0, w_{SR'} > 0,  \text{bias}_{R'} \ge -w_{SR'}\}$ for $\delta < 0$, or $\{w < 0, w_{SR'} > 0, \text{bias}_{R'} \ge -w_{SR'}\}$ for $\delta > 0$, will result in the global minimum loss of $2\sqrt{|\delta|}$. Note that non-zero initial weights will never cross over to another feasible region, as the nature of optimization guarantees $w \ne 0$ ($w=0$ results in the total loss blowing up to $\infty$) and $\text{bias}_{R'} \rightarrow 0$.\\

\noindent \textbf{Theorem 4.2}. Assume the same preconditions as Theorem 4.1, including assumptions (A1)-(A3) and (A6), but with non-sensitive attributes $\mathbf{X}$ and training with multiple ($k > 1$) representational disparity nodes. Assume that there is $\delta$-unfairness towards $S=1$ in the observed decision $O(\mathbf{X},S) =\text{Pr}_{\text{obs}}(H=1\:|\:\mathbf{X},S)$ for all $\mathbf{X=x}$, i.e., $\delta = \text{logit}(O(\mathbf{X},S=1))-\text{logit}(O(\mathbf{X},S=0)), \delta \ne 0$.

Assume that the outcome $Y$ does not depend on the sensitive attribute $S$, i.e., $\text{Pr}(Y=1\:|\:\mathbf{X}=\mathbf{x},S=s,H=h)=Y(\mathbf{x},h)$. Suppose the desired decision $D_{\mathbf{w}}(\textbf{x}, s) = \text{Pr}_\text{des}(H=1 \:|\: \mathbf{X}=\mathbf{x},S=s)$, parameterized by weight vector $\mathbf{w} = (\mathbf{w}_{\mathbf{X}R'_i}, w_{SR'_i}, w_i, \text{bias}_{R'_i})$, is learned using training data $T$ by minimizing the total loss $L_{\mathbf{w}}$,

\begin{align*}
&L_{\mathbf{w}} = aA_{\mathbf{w}} + bB_{\mathbf{w}} \\
&A_{\mathbf{w}} = \bigg|\sum_{\mathbf{x}}\text{Pr}(\mathbf{X=x})(Y(\mathbf{x},1)-Y(\mathbf{x},0))(D_{\mathbf{w}}(\mathbf{x},1) - D_{\mathbf{w}}(\mathbf{x},0))\bigg| \\  &B_{\mathbf{w}} = \sum_{i=1}^{k} (|w_i| + ||\mathbf{w}_{\mathbf{X}R'_i}||_{1} + |w_{SR'_i}| + |\text{bias}_{R'_i}|) \\ 
&\text{with,} \nonumber \\ &D_{\mathbf{w}}(\mathbf{x},s) = \sigma \left(\text{logit}(O(\mathbf{x},s)) + RD(\mathbf{x},s)\right) \\ &RD(\mathbf{x},s)=\sum_{i=1}^{k} RD_i(\mathbf{x},s) \\ &RD_i(\mathbf{x},s) = w_i\text{ReLU}(\mathbf{w}_{\mathbf{X}R'_i}\mathbf{x} + w_{SR'_i}s + \text{bias}_{R'_i})
\end{align*}

Here, $R'_1,.., R'_k$ are the representational disparity nodes used to explain the difference between the observed and desired human decision-maker; $RD_i(\mathbf{x},s)$ measures the representational disparity as captured by node $R_i$. Only $\mathbf{w}$ is updated while minimizing $L$. When $a \gg b$, the global minimum loss $L_{\text{min}}$ attained at $\mathbf{w}_{\text{min}}$ is,
\begin{align*}
&L_{\text{min}} = 2\sqrt{|\delta|} \\
&\mathbf{w}_{\text{min}} = \{\exists i \in \{1, ... , k\} ~\text{s.t.}, w_i=-\text{sign}(\delta)\sqrt{|\delta|}, \mathbf{w}_{\mathbf{X}R'_i}=0, w_{SR'_i}=\sqrt{|\delta|}, \text{bias}_{R'_i}=0, \nonumber \\ &~~~~~~~~~~~~~~~\forall j \neq i, j \in \{1, ... , k\}, w_j=0, \mathbf{w}_{\mathbf{X}R'_j}=0, w_{SR'_j}=0, \text{bias}_{R'_j}=0\}.
\end{align*}\\

\noindent \textit{Proof:} \noindent Since $a \gg b$, we minimize $B_\mathbf{w}$ under the constraint that $A_\mathbf{w} = 0$.  Let $c(\mathbf{x}) = 
\text{Pr}(\mathbf{X}=\mathbf{x})(Y(\mathbf{x},1)-Y(\mathbf{x},0))$, 
and enumerate all values $\mathbf{x}_1, \mathbf{x}_2, ....., \mathbf{x}_n$ for which $c(\mathbf{x}_i) \ne 0$. Now, we minimize $B_\mathbf{w}$ under the constraint that $A_\mathbf{w} = \sum_{i=1}^{n}c(\mathbf{x}_i)\bigg[D_{\mathbf{w}}(\mathbf{x}_i,1) -D_{\mathbf{w}}(\mathbf{x}_i,0)\bigg] = 0$. There are two cases:

\subsection*{I: $D_{\mathbf{w}}(\mathbf{x},1) = D_{\mathbf{w}}(\mathbf{x},0),~\forall \mathbf{x} \in \{\mathbf{x}_1, ..., \mathbf{x}_n\}$.}

\begin{align*}
    &~~~~~~~~~~~D_{\mathbf{w}}(\mathbf{x},1) = D_{\mathbf{w}}(\mathbf{x},0)\\ &\iff \sigma(\text{logit}(O(\mathbf{x},1)) + \sum_{i=1}^{k} ~w_i\text{ReLU}(\mathbf{w}^{T}_{\mathbf{X}R'_i}\mathbf{x} + w_{SR'_i} + \text{bias}_{R'_i})) \nonumber \\&~~~~~~=\sigma(\text{logit}(O(\mathbf{x},0)) + \sum_{i=1}^{k}~w_i\text{ReLU}(\mathbf{w}^{T}_{\mathbf{X}R'_i}\mathbf{x} + \text{bias}_{R'_i})) \\ &\iff \sum_{i=1}^{k}~w_i\bigg[\text{ReLU}(\mathbf{w}^{T}_{\mathbf{X}R'_i}\mathbf{x} + \text{bias}_{R'_i}) - \text{ReLU}(\mathbf{w}^{T}_{\mathbf{X}R'_i}\mathbf{x} + w_{SR'_i} + \text{bias}_{R'_i})\bigg] = \delta.
\end{align*}
\noindent Now, we find the $\mathbf{w}$ that minimizes $B_{\mathbf{w}}$,
\begin{align*}
    &\underset{\substack{w_1, w_{\mathbf{X}R'_1}, w_{SR'_1}, \text{bias}_{R'_1} \\ ,..., w_k, w_{\mathbf{X}R'_k}, w_{SR'_k}, \text{bias}_{R'_k}}}{\text{min}} ~\sum_{i=1}^{k}~|w_i| + ||\mathbf{w}_{\mathbf{X}R'_i}||_{1} + |w_{SR'_i}| + |\text{bias}_{R'_i}| \\
    &\text{s.t.,} ~\sum_{i=1}^{k}~w_i\bigg[\text{ReLU}(\mathbf{w}^{T}_{\mathbf{X}R'_i}\mathbf{x} + \text{bias}_{R'_i}) - \text{ReLU}(\mathbf{w}^{T}_{\mathbf{X}R'_i}\mathbf{x} + w_{SR'_i} + \text{bias}_{R'_i})\bigg] = \delta, ~\forall \mathbf{x} \in \{\mathbf{x}_1, ..., \mathbf{x}_n\}. 
\end{align*}
Let the contribution from node $R'_i$ to $\delta$ be $d_i\delta$. Then, the above optimization can be re-written as 
\begin{align*}
 &\underset{d_1,..,d_k}{\text{min}}\sum_{i=1}^{k}~\bigg[\underset{w_i, w_{\mathbf{X}R'_i}, w_{SR'_i}, \text{bias}_{R'_i}}{\text{min}} ~|w_i| + ||\mathbf{w}_{\mathbf{X}R'_i}||_{1} + |w_{SR'_i}| + |\text{bias}_{R'_i}|\bigg]\\
    &\text{s.t.,} ~w_i\bigg[\text{ReLU}(\mathbf{w}^{T}_{\mathbf{X}R'_i}\mathbf{x} + \text{bias}_{R'_i}) - \text{ReLU}(\mathbf{w}^{T}_{\mathbf{X}R'_i}\mathbf{x} + w_{SR'_i} + \text{bias}_{R'_i})\bigg] = d_i\delta, \\ &~~~~~~~d_1 + ... + d_k = 1, ~\forall i \in \{1,..,k\}, \forall \mathbf{x} \in \{\mathbf{x}_1, ... , \mathbf{x}_n\}. 
\end{align*}
\noindent Now, we find $\mathbf{w}$ that minimizes $|w_i| + ||\mathbf{w}_{XR'_i}||_{1} + |w_{SR'_i}| + |\text{bias}_{R'_i}|$, given the constraints, for given values of $d_1,\ldots,d_k$.
Using $|\text{ReLU}(a)-\text{ReLU}(b)| \leq |a-b|$, we obtain 
\begin{align*}
|w_iw_{SR'_i}| &= |w_i||(\mathbf{w}^{T}_{\mathbf{X}R'_i}\mathbf{x} + \text{bias}_{R'_i}) - (\mathbf{w}^{T}_{\mathbf{X}R'_i}\mathbf{x} + w_{SR'_i} + \text{bias}_{R'_i})| \\ &\ge |w_i||\text{ReLU}(\mathbf{w}^{T}_{\mathbf{X}R'_i}\mathbf{x} + \text{bias}_{R'_i}) - \text{ReLU}(\mathbf{w}^{T}_{\mathbf{X}R'_i}\mathbf{x} + w_{SR'_i} + \text{bias}_{R'_i})| \\ &= |w_i\text{ReLU}(\mathbf{w}^{T}_{\mathbf{X}R'_i}\mathbf{x} + \text{bias}_{R'_i}) - w_i\text{ReLU}(\mathbf{w}^{T}_{\mathbf{X}R'_i}\mathbf{x} + w_{SR'_i} + \text{bias}_{R'_i})| \\
&=|d_i\delta|.
\end{align*}
Hence,
\begin{align}
|w_i| + ||\mathbf{w}_{\mathbf{X}R'_i}||_{1} + |w_{SR'_i}| + |\text{bias}_{R'_i}| &\geq |w_i| + |w_{SR'_i}| \nonumber \\ &\geq 2\sqrt{|w_iw_{SR'_i}|} \label{amgm} \\ &\geq 2\sqrt{|d_i\delta|} \label{minbw}
\end{align}
Hence, minimum loss of $2\sqrt{|d_i\delta|}$ is obtained when $\mathbf{w}_{\mathbf{X}R'_i}=0$ and $\text{bias}_{R'_i}=0$. Eq. \ref{amgm} attains equality when $|w_i|=|w_{SR'_i}|$. Eq. \ref{minbw} attains equality when $|w_i|=|w_{SR'_i}|=\sqrt{|d_i\delta|}$.

%minimize first w.r.t to one set of variables, then the other set of variables https://math.stackexchange.com/questions/3843735/can-we-minimize-two-variable-function-by-minimizing-w-r-t-one-variable-first-fol
For a given $d_1,...,d_k$, the above optimization simplifies to
\begin{align*}
 &\underset{d_1,..,d_{k}}{\text{min}}~2\sqrt{|\delta|}[\sqrt{|d_1|} + ... + \sqrt{|d_{k-1}|} + \sqrt{|d_k|}] \\ &~~~~~~~\text{s.t.}~d_1 + ... + d_k = 1.
\end{align*}
For any setting of $d_1$, ... , $d_k$ (including the one minimizing the above loss),
\begin{align}
[\sqrt{|d_1|} + ... + \sqrt{|d_k|}]^2 &= |d_1| + ... + |d_{k}| + 2\sum_{i=1}^{k}\sum_{j=1}^{i-1} \sqrt{|d_{i}|}\sqrt{|d_{j}|} \nonumber \\ &\ge |d_1| + ... + |d_{k}| \label{eq10} \\ &\ge |d_1 + ... + d_{k}| = 1.\nonumber
\end{align}
Eq.~\ref{eq10} attains equality when $ 2\sum_{i=1}^{k}\sum_{j=1}^{i-1} \sqrt{|d_{i}|}\sqrt{|d_{j}|}=0$, i.e., when $\sqrt{|d_{i}|}\sqrt{|d_{j}|}=0 ~\forall i \neq j$. There exists at least one $d_i \neq 0$ as $d_1 + ... + d_k = 1$. Hence, $d_j=0$, $\forall j \neq i$ to make $\sqrt{|d_{i}|}\sqrt{|d_{j}|}=0 ~\forall i \neq j$. Therefore, exactly one $d_i \neq 0$. In other words, only one node out of $R'_1, ..., R'_k$ gets activated. Hence, the solution is,
\begin{align*}
    \mathbf{w}_{\text{min}} &= \{\exists i \in \{1, ... , k\} ~\text{s.t.}~w_i=-\text{sign}(\delta)\sqrt{|\delta|}, \mathbf{w}_{\mathbf{X}R'_i}=0, w_{SR'_i}=\sqrt{|\delta|}, \text{bias}_{R'_i}=0 \nonumber \\ &~~~~~~~w_j=0, \mathbf{w}_{\mathbf{X}R'_j}=0, w_{SR'_j}=0, \text{bias}_{R'_j}=0, \forall j \neq i, j \in \{1, ... , k\}\} \\ L_{\mathbf{w}_{\text{min}}} &= 2\sqrt{|\delta|}.
\end{align*} 

\subsection*{II: $~\exists \mathbf{x} \in \{\mathbf{x}_1, ..., \mathbf{x}_n\}$ s.t. $D_{\mathbf{w}}(\mathbf{x},1) \ne D_{\mathbf{w}}(\mathbf{x},0)$.}

Let $D_{\mathbf{w}}(\mathbf{x}_i,1) \ne D_{\mathbf{w}}(\mathbf{x}_i,0)$. Then, $\exists j \in \{1,...,n\}$, $j\ne i$, s.t., $D_{\mathbf{w}}(\mathbf{x}_j,1) \ne D_{\mathbf{w}}(\mathbf{x}_j,0)$, with
$\text{sign}(c(\mathbf{x}_i)(D_{\mathbf{w}}(\mathbf{x}_i,1) - D_{\mathbf{w}}(\mathbf{x}_i,0)))=-\text{sign}(c(\mathbf{x}_j)(D_{\mathbf{w}}(\mathbf{x}_j,1) - D_{\mathbf{w}}(\mathbf{x}_j,0)))$,
to satisfy the constraint $\sum_{i=1}^{n}c(\mathbf{x}_i)[D_{\mathbf{w}}(\mathbf{x}_i,1) -D_{\mathbf{w}}(\mathbf{x}_i,0)] = 0$. 

Without loss of generality, we assume that $c_i > 0$, $D_{\mathbf{w}}(\mathbf{x}_i,1) - D_{\mathbf{w}}(\mathbf{x}_i,0) > 0$, $c_j > 0$, and $D_{\mathbf{w}}(\mathbf{x}_j,1) - D_{\mathbf{w}}(\mathbf{x}_j,0) < 0$. Other feasible settings can be reduced to the above setting by performing one or more of the following operations: 
\begin{enumerate}
    \item Multiply $c_i$ by -1 and multiply $D_{\mathbf{w}}(\mathbf{x}_i, 1) - D_{\mathbf{w}}(\mathbf{x}_i, 0)$ by -1.
    \item Multiply $c_j$ by -1 and multiply $D_{\mathbf{w}}(\mathbf{x}_j, 1) - D_{\mathbf{w}}(\mathbf{x}_j, 0)$ by -1.
    \item Exchange $\mathbf{x}_i$ and $\mathbf{x}_j$.
\end{enumerate}

When $\delta < 0$, we show that the minimum loss attained with $D_{\mathbf{w}}(\mathbf{x}_i,1) - D_{\mathbf{w}}(\mathbf{x}_i,0) > 0$
is strictly greater than $2\sqrt{|\delta|}$.
\begin{align*}
&~~~~~~~~~~~D_{\mathbf{w}}(\mathbf{x}_i,1) - D_{\mathbf{w}}(\mathbf{x}_i,0) > 0 \\ &\iff \sum_{i=1}^{k}~w_i\bigg[\text{ReLU}(\mathbf{w}^{T}_{\mathbf{X}R'_i}\mathbf{x}_i + \text{bias}_{R'_i}) - \text{ReLU}(\mathbf{w}^{T}_{\mathbf{X}R'_i}\mathbf{x}_i + w_{SR'_i} + \text{bias}_{R'_i})\bigg] = \delta - \gamma, \gamma > 0.
\end{align*}
\noindent Now, we find $\mathbf{w}$ that minimizes the $B_{\mathbf{w}}$, $\forall \mathbf{x} \in \{\mathbf{x}_1, ..., \mathbf{x}_n\}, \gamma > 0$,
\begin{align*}
    &\underset{\substack{w_1, \mathbf{w}_{\mathbf{X}R'_1}, w_{SR'_1}, \text{bias}_{R'_1},..., \\ w_k, \mathbf{w}_{\mathbf{X}R'_k}, w_{SR'_k}, \text{bias}_{R'_k}, \gamma}}{\text{min}} ~\sum_{i=1}^{k}~|w_i| + ||\mathbf{w}_{\mathbf{X}R'_i}||_{1} + |w_{SR'_i}| + |\text{bias}_{R'_i}| \\
    &\text{s.t.,} ~\sum_{i=1}^{k}~w_i\bigg[\text{ReLU}(\mathbf{w}^{T}_{\mathbf{X}R'_i}\mathbf{x}_i + \text{bias}_{R'_i}) - \text{ReLU}(\mathbf{w}^{T}_{\mathbf{X}R'_i}\mathbf{x}_i + w_{SR'_i} + \text{bias}_{R'_i})\bigg] = \delta - \gamma.
\end{align*}
Let the contribution from $R'_i$ to $\delta$ be $d_i(\delta-\gamma)$. Then the above optimization can be written as 
\begin{align*}
 &\underset{d_1,..,d_k, \gamma}{\text{min}}\sum_{i=1}^{k}~\bigg[\underset{w_i, \mathbf{w}_{\mathbf{X}R'_i}, w_{SR'_i}}{\text{min}} ~|w_i| + ||\mathbf{w}_{\mathbf{X}R'_i}||_{1} + |w_{SR'_i}| + |\text{bias}_{R'_i}|\bigg]\\
    &\text{s.t.,} ~w_i\bigg[\text{ReLU}(\mathbf{w}^{T}_{\mathbf{X}R'_i}\mathbf{x} + \text{bias}_{R'_i}) - \text{ReLU}(\mathbf{w}^{T}_{\mathbf{X}R'_i}\mathbf{x} + w_{SR'_i} + \text{bias}_{R'_i})\bigg] = d_i(\delta-\gamma) \\ &~~~~~~~d_1 + ... + d_k = 1, ~\forall i \in \{1,..,k\}, \forall \mathbf{x} \in \{\mathbf{x}_1, ... , \mathbf{x}_n\}, \gamma > 0.
\end{align*}
%minimize first w.r.t to one set of variables, then the other set of variables https://math.stackexchange.com/questions/3843735/can-we-minimize-two-variable-function-by-minimizing-w-r-t-one-variable-first-fol
For a given $d_1,...,d_k$, following steps similar to Case I, the above optimization simplifies to,
\begin{align*}
 &\underset{d_1,..,d_{k}, \gamma}{\text{min}}~2\sqrt{|\delta|+|\gamma|}[\sqrt{|d_1|} + ... + \sqrt{|d_{k-1}|} + \sqrt{|d_k|}] \\ &~~~~~~~\text{s.t.}~d_1 + ... + d_k = 1 
\end{align*}
For any setting of $d_1$, ... , $d_k$ (including the one minimizing the above loss),
\begin{align}
[\sqrt{|d_1|} + ... + \sqrt{|d_{k}|} + \sqrt{|d_k|}]^2 &= |d_1| + ... + |d_{k}| + 2\sum_{i=1}^{i=k}\sum_{j=1}^{i-1}\sqrt{|d_{i}|}\sqrt{|d_{j}|} \nonumber \\ &\ge |d_1| + ... + |d_{k}| \label{eq11} \\ &\ge |d_1 + ... + d_{k}| = 1. \nonumber
\end{align}
Eq.~\ref{eq11} attains equality when $ 2\sum_{i=1}^{k}\sum_{j=1}^{i-1} \sqrt{|d_{i}|}\sqrt{|d_{j}|}=0$, i.e., when $\sqrt{|d_{i}|}\sqrt{|d_{j}|}=0 ~\forall i \neq j$. There exists at least one $d_i \neq 0$ as $d_1 + ... + d_k = 1$. Hence, $d_j=0$, $\forall j \neq i$ to make $\sqrt{|d_{i}|}\sqrt{|d_{j}|}=0 ~\forall i \neq j$. Therefore, exactly one $d_i \neq 0$. In other words, only one node out of $R'_1, ..., R'_k$ gets activated. The solution is
\begin{align*}
    \mathbf{w}_{\text{min}} &= \{\exists i \in \{1, ... , k\} ~\text{s.t.}~w_i=-\text{sign}(\delta)\sqrt{|\delta|}, \mathbf{w}_{\mathbf{X}R'_i}=0, w_{SR'_i}=\sqrt{|\delta|}, \text{bias}_{R'_i}=0 \nonumber \\ &~~~~~~~w_j=0, \mathbf{w}_{\mathbf{X}R'_j}=0, w_{SR'_j}=0, \text{bias}_{R'_j}=0, \forall j \neq i, j \in \{1, ... , k\}\} \\ L_{\mathbf{w}_{\text{min}}} &= 2\sqrt{|\delta|+|\gamma|}, \gamma > 0.
\end{align*} 

Similarly, when $\delta > 0$, one can show that the solution is
\begin{align*}
    \mathbf{w}_{\text{min}} &= \{\exists i \in \{1, ... , k\} ~\text{s.t.}~w_i=-\text{sign}(\delta)\sqrt{|\delta|}, \mathbf{w}_{\mathbf{X}R'_i}=0, w_{SR'_i}=\sqrt{|\delta|}, \text{bias}_{R'_i}=0 \nonumber \\ &~~~~~~~w_j=0, \mathbf{w}_{\mathbf{X}R'_j}=0, w_{SR'_j}=0, \text{bias}_{R'_j}=0, \forall j \neq i, j \in \{1, ... , k\}\} \\ L_{\mathbf{w}_{\text{min}}} &= 2\sqrt{|\delta|+|\gamma'|}, \gamma' > 0.
\end{align*} 
Hence, when $D_{\mathbf{w}}(\mathbf{x},1) \ne D_{\mathbf{w}}(\mathbf{x},0)$ for some $\mathbf{x} \in \{\mathbf{x}_1,..,\mathbf{x}_n\}$, the minimum loss obtained in any setting is strictly greater than $2\sqrt{|\delta|}$, which is the minimum loss obtained when $D_{\mathbf{w}}(\mathbf{x},1) = D_{\mathbf{w}}(\mathbf{x},0)$ for all $\mathbf{x} \in \{\mathbf{x}_1,..,\mathbf{x}_n\}$. $\blacksquare$\\

\noindent \textbf{Theorem 4.3} Assume the same preconditions as Theorem 4.1, including assumptions (A1)-(A5), but with disparity loss comparable to interpretability loss ($a \approx b$). Assume that there is $\delta$-unfairness towards $S=1$ in the observed decision $O(s) = \text{Pr}_{\text{obs}}(H=1\:|\:S=s)$, i.e., $\text{logit}(O(1))-\text{logit}(O(0)) = \delta$, $\delta \ne 0$. Assume that the only attribute is the sensitive attribute $S$ and that the outcome $Y$ depends only on the human decision $H$, i.e., $\text{Pr}
(Y=1\:|\:S=s,H=h)=Y(h)$, with $Y(1)-Y(0) = \alpha$. Suppose the desired decision $D_{\mathbf{w}}(s) = \text{Pr}_\text{des}(H=1 \:|\: S=s)$, parameterized by weight vector $\mathbf{w} = (w, w_{SR'},\text{bias}_{R'})$, is learned using training data $T$ by minimizing the total loss $L_{\mathbf{w}}$, where
\begin{align*}
&L_{\mathbf{w}} = aA_{\mathbf{w}} + bB_{\mathbf{w}}, ~0 < a < 1 \\
&A_{\mathbf{w}} = |\alpha||D_{\mathbf{w}}(1)-D_{\mathbf{w}}(0)| \\  &B_{\mathbf{w}} = |w| + |w_{SR'}| + |\text{bias}_{R'}| \\ &\text{with,} \nonumber \\ &D_{\mathbf{w}}(s) = \sigma \left(\text{logit}(O(s)) + RD(s)\right) \\ &RD(s) = w\text{ReLU}(w_{SR'}s + \text{bias}_{R'})
\end{align*}
Here, $R'$ is the disparity node; $RD(s)$ measures the representational disparity; $\mathbf{w}$ is comprised of $\{w_{SR'}, w, \text{bias}_{R'}\}$, in which $w_{SR'}$ is the weight from $S$ to $R'$, $w$ is the weight from $R'$ to $H$, and $\text{bias}_{R'}$ is the bias in $R'$. When $a + b = 1$, $a \approx b$, the global minimum loss $L_{\text{min}}$ attained at $\mathbf{w}_{\text{min}}$ is,
\begin{align*}
&L_{\text{min}} = \left\{\begin{array}{lr}
        \text{min}\{\underbrace{\underset{B_\mathbf{w}}{\text{min}}~(1-a)B_\mathbf{w} + a|\alpha|SD\left(\frac{B_\mathbf{w}^2}{4}\right)}_{L1}, ~\underbrace{\underset{B_\mathbf{w}}{\text{min}}~(1-a)B_\mathbf{w} + a|\alpha|EI\left(\frac{B_\mathbf{w}^2}{4}\right)\}}_{L2}, &~\delta > 0\\
        \text{min}\{\underbrace{\underset{B_\mathbf{w}}{\text{min}}~(1-a)B_\mathbf{w} + a|\alpha|SI\left(\frac{B_\mathbf{w}^2}{4}\right)}_{L3}, ~\underbrace{\underset{B_\mathbf{w}}{\text{min}}~(1-a)B_\mathbf{w} + a|\alpha|ED\left(\frac{B_\mathbf{w}^2}{4}\right)\}}_{L4} &~\delta < 0,
        \end{array}\right.
\end{align*}
and the weights learned corresponding to different losses are
\begin{align*}
\mathbf{w}_{\text{min}} &= \{w = -\frac{B_\text{opti}}{2}, w_{SR'} = \frac{B_\text{opti}}{2}, \text{bias}_{R'}=0\}, \text{when}~\text{loss L1 is chosen,} \\ \mathbf{w}_{\text{min}} &= \{w = \frac{B_\text{opti}}{2}, w_{SR'} = 0, \text{bias}_{R'}= \frac{B_\text{opti}}{2}\}, \text{when}~\text{loss L2 is chosen,}\\ \mathbf{w}_{\text{min}} &= \{w = \frac{B_\text{opti}}{2}, w_{SR'} = \frac{B_\text{opti}}{2}, \text{bias}_{R'}=0\}, \text{when}~\text{loss L3 is chosen, and}\\ \mathbf{w}_{\text{min}} &= \{w = - \frac{B_\text{opti}}{2}, w_{SR'} = 0, \text{bias}_{R'} = \frac{B_\text{opti}}{2}\}, \text{when}~\text{loss L4 is chosen,}
\end{align*}
where $B_\text{opti}$ is the optimal $B_\mathbf{w}$, $SD$ is a decrease in logit of the sigmoid with the larger logit, $EI$ is an equal increase in logit, $SI$ is an increase in logit of the sigmoid with the smaller logit, and $ED$ is an equal decrease in logit:
\begin{align*}
SD(x) &= \sigma \left(\text{logit}(O(0)) + \delta - x\right)-\sigma \left(\text{logit}(O(0))\right) \\ EI(x) &= \sigma \left(\text{logit}(O(0)) + \delta + x\right) - \sigma \left(\text{logit}(O(0)) + x\right) \\ SI(x) &= \sigma \left(\text{logit}(O(0))\right) -\sigma \left(\text{logit}(O(0)) + \delta + x\right) \\ ED(x) &= \sigma \left(\text{logit}(O(0)) - x\right)-\sigma \left(\text{logit}(O(0)) + \delta - x\right).
\end{align*}
\textit{Proof:}
\begin{align*}
&~~~~~\underset{\mathbf{w}}{\text{min}} ~[aA_\mathbf{w} + (1-a)B_\mathbf{w}] \\ &= \underset{B_\mathbf{w}}{\text{min}} ~\underset{\substack{\mathbf{w}: |w| + |w_{SR'}| + |\text{bias}_{R'}|=B_\mathbf{w}}}{\text{min}} ~[aA_\mathbf{w} + (1-a)B_\mathbf{w}] \\ &= \underset{B_\mathbf{w}}{\text{min}} ~[(1-a)B_\mathbf{w} ~+~ \underset{\substack{\mathbf{w}: |w| + |w_{SR'}| + |\text{bias}_{R'}|=B_\mathbf{w}}}{\text{min}} a|\alpha||\sigma \left(\text{logit}(O(1)) + w\text{ReLU}(w_{SR'} + \text{bias}_{R'})\right) \nonumber \\ &~~~~~~~~~~~~- \sigma \left(\text{logit}(O(0)) + w\text{ReLU}(\text{bias}_{R'})\right)|] \\ &= \underset{B_\mathbf{w}}{\text{min}} ~[(1-a)B_\mathbf{w} ~+~ \underset{\substack{\mathbf{w}: |w| + |w_{SR'}| + |\text{bias}_{R'}|=B_\mathbf{w}}}{\text{min}} a|\alpha||\sigma \left(\text{logit}(O(0)) + \delta + w\text{ReLU}(w_{SR'} + \text{bias}_{R'})\right) \nonumber \\ &~~~~~~~~~~~~-\sigma \left(\text{logit}(O(0)) + w\text{ReLU}(\text{bias}_{R'})\right)|].
\end{align*}
Note that $w_{SR'} + \text{bias}_{R'} \ge 0$ and $\text{bias}_{R'} \ge 0$. To see this, suppose $\text{bias}_{R'} < 0$. Then adding
$\text{bias}_{R'}$ to $w_{SR'}$ and setting $\text{bias}_{R'}$ to 0 results in the same $A_\mathbf{w}$ with reduced $B_\mathbf{w}$. Hence, $\text{bias}_{R'} \ge 0$. Similarly, suppose $w_{SR'} + \text{bias}_{R'} < 0$ and $\text{bias}_{R'} \ge 0$. Then setting $w_{SR'}$ to $-\text{bias}_{R'}$ results in the same $A_\mathbf{w}$ with reduced $B_\mathbf{w}$. Hence, $w_{SR'} + \text{bias}_{R'} \ge 0$. 

First, we solve the inner optimization problem. 
\begin{align}
&\underset{\substack{\mathbf{w}: |w| + |w_{SR'}| + |\text{bias}_{R'}|=B_\mathbf{w} \\ |w|[|w_{SR'}| + |\text{bias}_{R'}|] = c}}{\text{min}} ~\underset{\substack{0 \leq k \leq 1 \\ s_1 = \pm 1, \\ s_2 = \pm 1}}{\text{min}}|\sigma \left(\text{logit}(O(0)) + \delta + s_1(1-k)c + s_2kc\right)-\sigma \left(\text{logit}(O(0)) + s_2kc\right)|,  \label{ref_eq}
\end{align}
where $k$ is the fraction of $c$ assigned to $|w\text{bias}_{R'}|$ and $1-k$ is the fraction assigned to $|ww_{SR'}|$, i.e., 
\begin{align}
kc &= |w\text{bias}_{R'}| \label{usageofkfrac0} \\ (1-k)c &= |ww_{SR'}| \label{usageofkfrac1}
\end{align}
and $s_1$ is the sign of $ww_{SR'}$ and $s_2$ is the sign of $w\text{bias}_{R'}$. Note that the inner optimization in Eq. \ref{ref_eq} depends only on $c, k, s_1$ and $s_2$. Hence, out of all the $B_\mathbf{w}$ satisfying Eq. \ref{ref_eq} for a given $c$, we choose the minimum $B_\mathbf{w}$.  

Let us find the minimum $B_\mathbf{w}$ for a given $c$.
\begin{align*}
\underset{w}{\text{min}}~|w|+\frac{c}{|w|} = 2\sqrt{c} = B_\mathbf{w} ~\text{or}~ c = \frac{B_\mathbf{w}^2}{4}
\end{align*}
The inner optimization problem is,
\begin{align*}
&~~~~~\underset{\substack{0 \leq k \leq 1, \\ s_1 = \pm 1, \\ s_2 = \pm 1}}{\text{min}}\bigg|\sigma \left(\text{logit}(O(0)) + \delta + s_1(1-k)\frac{B_\mathbf{w}^2}{4}+s_2k\frac{B_\mathbf{w}^2}{4}\right)-\sigma \left(\text{logit}(O(0)) + s_2k\frac{B_\mathbf{w}^2}{4}\right)\bigg| \\ &~~~~~=\text{min}\bigg\{\text{Loss $A1$}, \text{Loss $A2$}, \text{Loss $A3$}, \text{Loss $A4$}\bigg\},
\end{align*}
where Loss $A1$ is obtained by setting $s_1=1$ and $s_2=1$; Loss $A2$ is obtained by setting $s_1=1$ and $s_2=-1$; Loss $A3$ is obtained by setting $s_1=-1$ and $s_2=1$; and Loss $A4$ is obtained by setting $s_1=-1$ and $s_2=-1$.

\textbf{Loss $A1$:}\\
When $\delta > 0$,
\begin{align*}
&\underset{0 \leq k \leq 1}{\text{min}}\bigg|\sigma \left(\text{logit}(O(0)) + \delta + \frac{B_\mathbf{w}^2}{4}\right)-\sigma \left(\text{logit}(O(0)) + k\frac{B_\mathbf{w}^2}{4}\right)\bigg| \\ &~~~~~=\sigma \left(\text{logit}(O(0)) + \delta + \frac{B_\mathbf{w}^2}{4}\right)-\sigma \left(\text{logit}(O(0)) + \frac{B_\mathbf{w}^2}{4}\right),
\end{align*}
as $\sigma$ is an increasing function and $\text{logit}(O(0)) + \delta + \frac{B_\mathbf{w}^2}{4} \geq \text{logit}(O(0)) + \frac{B_\mathbf{w}^2}{4}$ when $\delta > 0$. In this case, the logit in both sigmoids is increased to decrease the sigmoid difference. 

When $\delta < 0$,
\begin{align*}
&~~~~~\underset{0 \leq k \leq 1}{\text{min}}\bigg|\sigma \left(\text{logit}(O(0)) + \delta + \frac{B_\mathbf{w}^2}{4}\right)-\sigma \left(\text{logit}(O(0)) + k\frac{B_\mathbf{w}^2}{4}\right)\bigg| \\ &=\underset{0 \leq k \leq 1}{\text{min}}\bigg|\sigma \left(\text{logit}(O(0)) + k\frac{B_\mathbf{w}^2}{4}\right) - \sigma \left(\text{logit}(O(0)) + \delta + \frac{B_\mathbf{w}^2}{4}\right)\bigg| \\ &~~~~~~~~~~=\bigg|\sigma \left(\text{logit}(O(0))\right)-\sigma \left(\text{logit}(O(0)) + \delta + \frac{B_\mathbf{w}^2}{4}\right)\bigg| \\ &~~~~~~~~~~=\sigma \left(\text{logit}(O(0))\right)-\sigma \left(\text{logit}(O(0)) + \delta + \frac{B_\mathbf{w}^2}{4}\right),
\end{align*}
as $B_\mathbf{w} \leq 2\sqrt{|\delta|}$ and $\delta + \frac{B_\mathbf{w}^2}{4} \leq 0$. In this case, the logit in one sigmoid is decreased to decrease the sigmoid difference. Note that $B_\mathbf{w} > 2\sqrt{|\delta|}$ is not a feasible set as there exists a solution with $A_\mathbf{w}=0$ and $B_\mathbf{w}=2\sqrt{|\delta|}$ (We can set $A_\mathbf{w}=0$ and show using the same proof as in \textbf{Theorem 4.1} that the minimum $B_\mathbf{w}$ among the weights that make $A_\mathbf{w}=0$ is $B_\mathbf{w}=2\sqrt{|\delta|}$.)

\textbf{Loss $A2$:}\\
When $\delta > 0$,
\begin{align*}
&~~~~~~~~~~~~~~~~~\underset{0 \leq k \leq 1}{\text{min}}\bigg|\sigma \left(\text{logit}(O(0)) + \delta + (1-2k)\frac{B_\mathbf{w}^2}{4}\right)-\sigma \left(\text{logit}(O(0)) - k\frac{B_\mathbf{w}^2}{4}\right)\bigg| \\
&~~~~~~~~~~~~=\underset{0 \leq k \leq 1}{\text{min}}~\sigma \left(\text{logit}(O(0)) + \delta + (1-2k)\frac{B_\mathbf{w}^2}{4}\right)-\sigma \left(\text{logit}(O(0)) - k\frac{B_\mathbf{w}^2}{4}\right),
\end{align*}
as $\sigma$ is an increasing function and $\text{logit}(O(0)) + \delta + (1-2k)\frac{B_\mathbf{w}^2}{4} \geq \text{logit}(O(0)) - k\frac{B_\mathbf{w}^2}{4} ~\forall k \in [0,1]$ because $\text{logit}(O(0)) + \delta + (1-2k)\frac{B_\mathbf{w}^2}{4} - \text{logit}(O(0)) + k\frac{B_\mathbf{w}^2}{4} = \delta + (1-k)\frac{B_\mathbf{w}^2}{4} \geq 0$.

Let 
\begin{align*}
f(k) = \sigma \left(\text{logit}(O(0)) + \delta + (1-2k)\frac{B_\mathbf{w}^2}{4}\right) - \sigma \left(\text{logit}(O(0)) - k\frac{B_\mathbf{w}^2}{4}\right).
\end{align*}

Now we will show that $f''(k) < 0$ at $k$ where $f'(k)=0$. Consequently, we will show that $f(k)$ attains its minimum at $k=0$ and $k=1$. 

Let $x1 = \text{logit}(O(0)) + \delta$ and $x0 = \text{logit}(O(0))$. Now,
\begin{align*}
&~~~~~~f'(k) = 0 \equiv 2\sigma'\left(x1 + (1-2k)\frac{B_\mathbf{w}^2}{4}\right) = \sigma'\left(x0 - k\frac{B_\mathbf{w}^2}{4}\right).
\end{align*}
Since the maximum value of $\sigma'(x)$ is 1/4, $\sigma'\left(x1 + (1-2k) \frac{B_\mathbf{w}^2}{4}\right) \leq 1/8$ for the aforementioned equation to have a solution. This will only be the case for $|x1 + (1-2k)\frac{B_\mathbf{w}^2}{4}| \geq \ln(3+2\sqrt{2}) \approx 1.76$ (Note that for $y = \sigma'(x) = \sigma(x)(1-\sigma(x))$, $x = \pm \ln((1+\sqrt{1-4y})/(1-\sqrt{1-4y}))$). 

Further, $\sigma'(x)$ is an increasing function for $x < 0$, and $x0 - k\frac{B_\mathbf{w}^2}{4} \leq x1 + (1-2k)\frac{B_\mathbf{w}^2}{4} ~\forall k \in [0,1]$ as $\delta > 0$. Hence, $2\sigma'\left(x1 + (1-2k)\frac{B_\mathbf{w}^2}{4}\right) = \sigma'\left(x0 - k\frac{B_\mathbf{w}^2}{4}\right)$ cannot have a solution when $x1 + (1-2k)\frac{B_\mathbf{w}^2}{4} < 0$. Consequently, out of $|x1+(1-2k)\frac{B_\mathbf{w}^2}{4}| \geq \ln(3+2\sqrt{2}) \approx 1.76$, only $x1+(1-2k)\frac{B_\mathbf{w}^2}{4} \geq \ln(3+2\sqrt{2})$ needs to be considered for analyzing $f'(k)=0$. 

Now, we look at the second derivative,
\begin{align*}
f''(k) &=-4\frac{B_\mathbf{w}^4}{16}\sigma'\left(x1 + (1-2k)\frac{B_\mathbf{w}^2}{4}\right)g\left(x1 + (1-2k)\frac{B_\mathbf{w}^2}{4}\right) \nonumber \\ &~~~~~+ \frac{B_\mathbf{w}^4}{16}\sigma'\left(x0-k\frac{B_\mathbf{w}^2}{4}\right)g\left(x0-k\frac{B_\mathbf{w}^2}{4}\right),
\end{align*}
where $g(x) = (e^{x}-1)/(e^{x}+1)$, which is an increasing function of x. When the first derivative is 0, we can substitute $\sigma'\left(x0-k\frac{B_\mathbf{w}^2}{4}\right) = 2\sigma'\left(x1 + (1-2k)\frac{B_\mathbf{w}^2}{4}\right)$, and thus the second derivative is
\begin{align*}
f''(k) &= -4\frac{B_\mathbf{w}^4}{16}\sigma'\left(x1 + (1-2k)\frac{B_\mathbf{w}^2}{4}\right)g\left(x1 + (1-2k)\frac{B_\mathbf{w}^2}{4}\right) \nonumber \\ &~~~~+ 2\frac{B_\mathbf{w}^4}{16}\sigma'\left(x1 + (1-2k)\frac{B_\mathbf{w}^2}{4}\right)g\left(x0-k\frac{B_\mathbf{w}^2}{4}\right).
\end{align*}
Since $f$ is an increasing function and $x1 + (1-2k)\frac{B_\mathbf{w}^2}{4} \geq x0 - k\frac{B_\mathbf{w}^2}{4}$, 
\begin{align}
g\left(x0 - k\frac{B_\mathbf{w}^2}{4}\right) \leq g\left(x1 + (1-2k) \frac{B_\mathbf{w}^2}{4}\right). \label{cond1thm3}
\end{align}
Also, since $x1+(1-2k)\frac{B_\mathbf{w}^2}{4} \geq \ln(3+2\sqrt{2}) \approx 1.76$ and $f(x) > 0 ~\forall x > 0$, 
\begin{align}
g\left(x1+(1-2k)\frac{B_\mathbf{w}^2}{4}\right) > 0. \label{cond2thm3}
\end{align}

Using Eq. \ref{cond1thm3} and \ref{cond2thm3}, one can easily show that $f''(k) \leq 0$. Hence, $f(k)$ attains a local maximum when $f'(k)=0$. In other words, $\sigma \left(x1 + (1-2k)\frac{B_\mathbf{w}^2}{4}\right)-\sigma \left(x0 - k\frac{B_\mathbf{w}^2}{4}\right)$ attains the minimum at $k=0$ or $k=1$ as $f$ is a continuous function.

Therefore, when $\delta > 0$,
\begin{align*}
&~~~~\underset{0 \leq k \leq 1}{\text{min}}~\bigg|\sigma \left(\text{logit}(O(0)) + \delta + (1-2k)\frac{B_\mathbf{w}^2}{4}\right)-\sigma \left(\text{logit}(O(0)) - k\frac{B_\mathbf{w}^2}{4}\right)\bigg| \\ &=\text{min}~\bigg\{\sigma \left(\text{logit}(O(0)) + \delta + \frac{B_\mathbf{w}^2}{4}\right)-\sigma \left(\text{logit}(O(0))\right), \nonumber \\ &~~~~~~~~~~~~~~~~\sigma \left(\text{logit}(O(0)) + \delta - \frac{B_\mathbf{w}^2}{4}\right)-\sigma \left(\text{logit}(O(0)) - \frac{B_\mathbf{w}^2}{4}\right) \bigg\}.
\end{align*}
In this case, either the logit of one sigmoid is further increased, or the logit of both sigmoids is decreased to reduce the existing $\delta$ difference. 

Both of these cases are not optimal. The former case would have been optimal if the logit of one sigmoid is decreased to reduce the existing $\delta$ difference in the logits, and the latter case would have been optimal if only the logit of the sigmoid is decreased to reduce the existing $\delta$ difference in the logits. 

Similarly, when $\delta < 0$, one can show that
\begin{align*}
&~~~~\underset{0 \leq k \leq 1}{\text{min}}~\bigg|\sigma \left(\text{logit}(O(0)) + \delta + (1-2k)\frac{B_\mathbf{w}^2}{4}\right)-\sigma \left(\text{logit}(O(0)) - k\frac{B_\mathbf{w}^2}{4}\right)\bigg| \\ &=\text{min}~\bigg\{\sigma \left(\text{logit}(O(0))\right)-\sigma \left(\text{logit}(O(0)) + \delta + \frac{B_\mathbf{w}^2}{4}\right), \nonumber \\ &~~~~~~~~~~~~~~~~\sigma \left(\text{logit}(O(0)) - \frac{B_\mathbf{w}^2}{4}\right)-\sigma \left(\text{logit}(O(0)) + \delta - \frac{B_\mathbf{w}^2}{4}\right)\bigg\}.
\end{align*}
In this case, either the logit of one sigmoid is further increased or the logits of both the sigmoids are decreased to reduce the existing $\delta$ difference.

\textbf{Loss $A3$:}\\
Following a similar proof structure as for Loss $A2$, one can show that, when $\delta > 0$,
\begin{align*}
&~~~~\underset{0 \leq k \leq 1}{\text{min}}~\bigg|\sigma \left(\text{logit}(O(0)) + \delta + (2k-1)\frac{B_\mathbf{w}^2}{4}\right)-\sigma \left(\text{logit}(O(0)) + k\frac{B_\mathbf{w}^2}{4}\right)\bigg| \\ &=\text{min}~\bigg\{\sigma \left(\text{logit}(O(0)) + \delta - \frac{B_\mathbf{w}^2}{4}\right)-\sigma \left(\text{logit}(O(0))\right), \nonumber \\ &~~~~~~~~~~~~~~~~\sigma \left(\text{logit}(O(0)) + \delta + \frac{B_\mathbf{w}^2}{4}\right)-\sigma \left(\text{logit}(O(0)) + \frac{B_\mathbf{w}^2}{4}\right) \bigg\}.
\end{align*}
In this case, either the logit of one sigmoid is decreased to decrease the sigmoid difference, or the logit of both sigmoids are pushed to the extreme to decrease the sigmoid difference.

Similarly, when $\delta < 0$, one can show that,
\begin{align*}
&~~~\underset{0 \leq k \leq 1}{\text{min}}~\bigg|\sigma \left(\text{logit}(O(0)) + \delta + (2k-1)\frac{B_\mathbf{w}^2}{4}\right)-\sigma \left(\text{logit}(O(0)) + k\frac{B_\mathbf{w}^2}{4}\right)\bigg| \\ &=\text{min}~\bigg\{\sigma \left(\text{logit}(O(0))\right)-\sigma \left(\text{logit}(O(0)) + \delta - \frac{B_\mathbf{w}^2}{4}\right), \nonumber \\ &~~~~~~~~~~~~~~~\sigma \left(\text{logit}(O(0)) + \frac{B_\mathbf{w}^2}{4}\right)-\sigma \left(\text{logit}(O(0)) + \delta + \frac{B_\mathbf{w}^2}{4}\right)\bigg\}.
\end{align*}
In this case, either the logit of one sigmoid is decreased, or the logit of one sigmoid is increased while the logit of other sigmoid is increased to reduce the existing $\delta$ difference in the logits. 

Both of these cases are not optimal. The former case would have been optimal if the logit of one sigmoid is increased to reduce the existing $\delta$ difference in the logits, and the latter case would have been optimal if only the logit of one of the sigmoids is increased to reduce the existing $\delta$ difference in the logits.

\textbf{Loss $A4$:}\\
When $\delta > 0$,
\begin{align*}
&\underset{0 \leq k \leq 1}{\text{min}}\bigg|\sigma \left(\text{logit}(O(0)) + \delta - \frac{B_\mathbf{w}^2}{4}\right)-\sigma \left(\text{logit}(O(0)) - k\frac{B_\mathbf{w}^2}{4}\right)\bigg| \\ &~~~~~=\sigma \left(\text{logit}(O(0)) + \delta - \frac{B_\mathbf{w}^2}{4}\right)-\sigma \left(\text{logit}(O(0))\right).
\end{align*}
In this case, the logit of one sigmoid is decreased to reduce the existing $\delta$ difference in the logits.

When $\delta < 0$,
\begin{align*}
&~~~~~\underset{0 \leq k \leq 1}{\text{min}}\bigg|\sigma \left(\text{logit}(O(0)) + \delta - \frac{B_\mathbf{w}^2}{4}\right)-\sigma \left(\text{logit}(O(0)) - k\frac{B_\mathbf{w}^2}{4}\right)\bigg| \\ &~~~~~~~~~~=\sigma \left(\text{logit}(O(0)) - \frac{B_\mathbf{w}^2}{4}\right)-\sigma \left(\text{logit}(O(0)) + \delta - \frac{B_\mathbf{w}^2}{4}\right).
\end{align*}
In this case, the logit of both sigmoids is decreased to reduce the existing $\delta$ difference in the logits. 

\textbf{Combining all of the above solutions}, we can write the minimum loss as
\begin{align}
&\text{min}~\bigg\{\underset{B_\mathbf{w}}{\text{min}}~(1-a)B_\mathbf{w} + a|\alpha|\sigma \left(\text{logit}(O(0)) + \delta - \frac{B_\mathbf{w}^2}{4}\right)- a|\alpha|\sigma \left(\text{logit}(O(0))\right), \label{lossa} \\ &~~~~~~~~~~\underset{B_\mathbf{w}}{\text{min}}~(1-a)B_\mathbf{w} + a|\alpha|\sigma \left(\text{logit}(O(0)) + \delta + \frac{B_\mathbf{w}^2}{4}\right)-a|\alpha|\sigma \left(\text{logit}(O(0)) + \frac{B_\mathbf{w}^2}{4}\right) \bigg\}, \delta > 0, \label{lossb} \\ &\text{min}~\bigg\{\underset{B_\mathbf{w}}{\text{min}}~(1-a)B_\mathbf{w} + a|\alpha|\sigma \left(\text{logit}(O(0))\right)-a|\alpha|\sigma \left(\text{logit}(O(0)) + \delta + \frac{B_\mathbf{w}^2}{4}\right),\label{lossc}  \\ &~~~~~~~~~~\underset{B_\mathbf{w}}{\text{min}}~(1-a)B_\mathbf{w} + a|\alpha|\sigma \left(\text{logit}(O(0)) - \frac{B_\mathbf{w}^2}{4}\right)-a|\alpha|\sigma \left(\text{logit}(O(0)) + \delta - \frac{B_\mathbf{w}^2}{4}\right)\bigg\}, \delta < 0. \label{lossd}
\end{align}
Let us denote loss in Eq. \ref{lossa} as loss $L1$, loss in Eq. \ref{lossb} as loss $L2$, loss in Eq. \ref{lossc} as loss $L3$, and loss in Eq. \ref{lossd} as loss $L4$. We derive the optimal weights learned when different losses are chosen. Let $B_\text{opti}$ be the optimal $B_\mathbf{w}$ minimizing all the losses. Suppose that loss $L1$ is chosen. Then, either loss $A3$ or loss $A4$ could result in loss $L1$. Hence, the parameters could be $\{s_1=-1, s_2=1, k=0\}$ or $\{s_1=-1, s_2=-1, k=0\}$. Let $s_1=-1, s_2=1, k=0$ or $s_1=-1, s_2=-1, k=0$. Using Eq. \ref{usageofkfrac0} and \ref{usageofkfrac1},
\begin{align*}
    &ww_{SR'} = -\frac{B^2_\text{opti}}{4}, -w + w_{SR'} = B_\text{opti}, -w - \frac{B^2_\text{opti}}{4w} = B_\text{opti}, \\
    &\mathbf{w}_{\text{min}} = \{w = -\frac{B_\text{opti}}{2}, w_{SR'} = \frac{B_\text{opti}}{2}, \text{bias}_{R'}=0\}.
\end{align*}

Similarly, we can calculate the optimal weights when loss $L2$, $L3$, or $L4$ is selected, with optimal weights being $\{w=\frac{B_\text{opti}}{2}, w_{SR'} = 0, \text{bias}_{R'} = \frac{B_\text{opti}}{2}\}$, $\{w=\frac{B_\text{opti}}{2}, w_{SR'} = \frac{B_\text{opti}}{2}, \text{bias}_{R'}=0\}$, and 
$\{w=-\frac{B_\text{opti}}{2}, w_{SR'} = 0, \text{bias}_{R'} = \frac{B_\text{opti}}{2}\}$, respectively. $\blacksquare$ 

We present different settings of hyperparameters that result in optimal losses $L1$, $L2$, $L3$, and $L4$:
\begin{enumerate}
\item Loss $L1$ is globally optimal when $a=0.9, \text{logit}(O(0))=-4.595, \delta=5, \alpha=1$ as loss $L1 = 0.4$ and loss $L2 = 0.531$. Here, $B_\text{opti}=3.47$, $w=-1.735$, $w_{SR'}=1.735$, $\text{bias}_{R'}=0$. 
\item Loss $L2$ is globally optimal when $a=0.9, \text{logit}(O(0))=-2, \delta=10, \alpha=1$ as loss $L1 = 0.597$ and loss $L2 = 0.49$. Here, $B_\text{opti}=4.418$, $w=2.209$, $w_{SR'}=0$, $\text{bias}_{R'}=2.209$. 
\item Loss $L3$ is globally optimal when $a=0.9, \text{logit}(O(0))=4.595, \delta=-5, \alpha=1$ as loss $L3 = 0.4$ and loss $L4 = 0.531$. Here, $B_\text{opti}=3.47$, $w=1.735$, $w_{SR'}=1.735$, $\text{bias}_{R'}=0$. 
\item Loss $L4$ is globally optimal when $a=0.9, \text{logit}(O(0))=2, \delta=-10, \alpha=1$ as loss $L3 = 0.597$ and loss $L4 = 0.49$. Here, $B_\text{opti}=4.418$, $w=-2.209$, $w_{SR'}=0$, $\text{bias}_{R'}=2.209$. \label{hyperparamthm3}
\end{enumerate} 

\section{Experiments}
\label{expdetails}
In Appendix~\ref{expsynthetic}, we present several proof-of-concept experiments on simple, synthetic datasets, in support of our theoretical results in Section~\ref{sec:theory}.  In Appendix~\ref{expreal}, we present additional details for our real-world experiments in Section~\ref{exp_sec}.

\subsection{Experiments on Synthetic Data}
\label{expsynthetic}
\subsubsection{Validation of Theorem 4.2}
\textbf{Data Generation Process}\\
We generate a simple, synthetic dataset of size 100,000, split into 70,000 training records and 30,000 test records.  The dataset is generated to satisfy the preconditions of Theorem 4.2: the non-sensitive attributes $\mathbf{X}$ are independent of the sensitive attribute $S$, and the outcome $Y$ is conditionally independent of $S$ given the human decision $H$.  Moreover, there is $\delta$-unfairness toward $S=1$ in the observed decision $H$, i.e., for all $\mathbf{X}=\mathbf{x}$, we have $\text{logit}(H=1\:|\: \mathbf{X=x}, S=1)-\mbox{logit}(H=1\:|\:\mathbf{X=x},S=0) = \delta$. We assume a single non-sensitive attribute $X$, and assume that $X$, $S$, $H$, and $Y$ are all binary.  We generate $S \sim \text{Bernoulli}(0.5)$ and $X \sim \text{Bernoulli}(0.5)$. For the human decision, we generate $H \sim \text{Bernoulli}(0.6)$ for $S=0$, and $H \sim \text{Bernoulli}(0.3)$ for $S=1$.  This models a scenario where the protected class $S=1$ is less likely to receive the positive human decision $H=1$.  Finally, for the outcome, we generate $Y \sim \text{Bernoulli}(0.3)$ for ($X$, $H$) = (0, 0),
$Y \sim \text{Bernoulli}(0.8)$ for ($X$, $H$) = (0, 1), $Y \sim \text{Bernoulli}(0.2)$ for ($X$, $H$) = (1, 0), and $Y \sim \text{Bernoulli}(0.6)$ for ($X$, $H$) = (1, 1), regardless of the value of $S$.  This models a scenario where the positive human decision $H=1$ makes the positive outcome $Y=1$ more likely.

We note that the optimal cross-entropy loss for predicting the observed human decision $H$, given this data generating process, is $C_{opt} \approx 0.642$.  This can be easily computed as the expectation (over $X$ and $S$) of $- p_{xs} \log p_{xs} - (1-p_{xs}) \log (1-p_{xs})$, where $p_{xs} = \text{Pr}(H=1 \:|\: X=x, S=s)$.
Similarly, the optimal cross-entropy loss for predicting the outcome $Y$, given this data generating process, is $D_{opt} \approx 0.570$.
This can be easily computed as the expectation (over $X$, $S$, and $H$) of $-p_{xsh} \log p_{xsh} - (1-p_{xsh}) \log (1-p_{xsh})$, where $p_{xsh} = \text{Pr}(Y=1 \:|\: X=x, S=s, H=h)$.

\textbf{Training}\\
We use the four-layer neural network architecture described in Section~\ref{ourmodel}.  The first layer consists of inputs $S$ and $X$.  The second layer consists of three nodes $\mathbf{R}=\{R_1, R_2, R_3\}$ capturing the internal representation of $S$ and $X$. 
The third and fourth layers represent the human decision $H$ and outcome $Y$, as above.  The observed and desired decision-makers differ in the internal representations of the input used to make the human decision $H$. We assume that the observed decision-maker uses only $\{R_1\}$, and the desired decision-maker uses $\{R_1, R_2, R_3\}$.  We use Adam optimizer~\cite{kingma2014adam} to minimize Eq.~\ref{totalloss1}, with hyperparameters $a=0.999$, $b=0.001$, $c=1000$, and $d=1000$.  (Note that $a \approx 1$ is a precondition for Theorem 4.2.)  We train the neural network on the training data for 1000 epochs for each fit, and average the results across 100 fits. 

\textbf{Results}\\
For all 100 fits, we observe that the disparity in fairness loss $A$ converges to a value very close to zero ($\mathbf{O}(10^{-3})$), while the losses $C$ and $D$ (for modeling the observed human decision $H$ and the outcome $Y$ respectively) are very close to their optimal values $C_{opt}$ and $D_{opt}$ respectively.  Moreover, we observe that only one of the two representational disparity nodes ($R_2$ or $R_3$) has non-zero weights $w_{SR_j}$ and $w_{R_j}$ for a given fit, with 
$w_{SR_j}w_{R_j} \approx \delta$, where $\delta = \text{logit}(0.6)-\text{logit}(0.3) \approx 1.25$.  The other representational disparity node has $w_{SR_i}$ and $w_{R_i}$ very close to zero ($\mathbf{O}(10^{-3})$). These results demonstrate that the network converges to the globally optimal loss given in Theorem 4.2; we see that, even though convergence to these weights is not guaranteed, it is consistently achieved in practice.

\begin{figure}[t]
    \centering
    \subfigure[]{\includegraphics[width=0.21\textwidth]{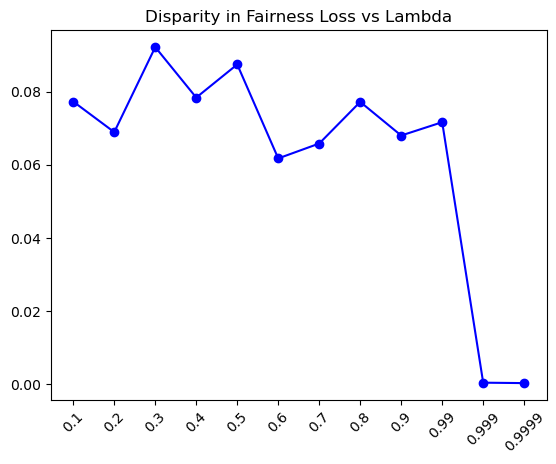}} 
    \subfigure[]{\includegraphics[width=0.2\textwidth]{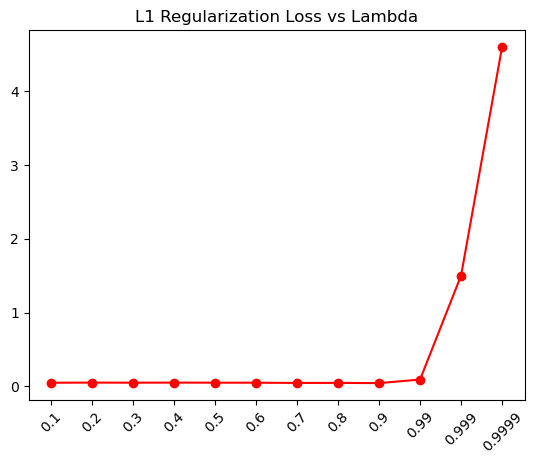}}
    \subfigure[]{\includegraphics[width=0.21\textwidth]{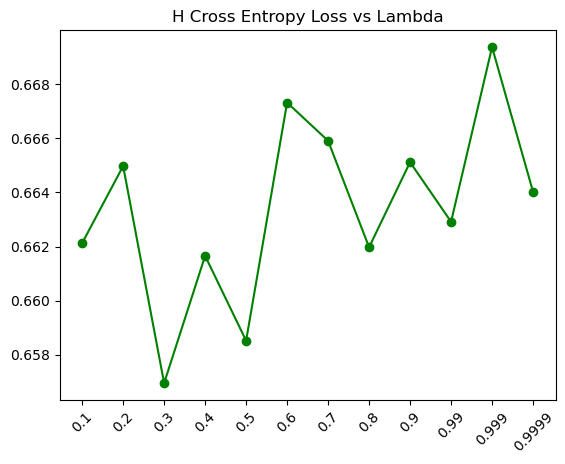}} 
    \subfigure[]{\includegraphics[width=0.21\textwidth]{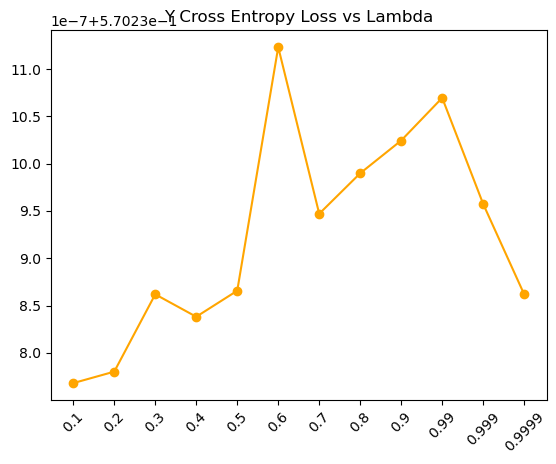}}
    \caption{Components of the multi-objective loss function as a function of $a$, the relative weight of the disparity in fairness loss as compared to the regularization loss. Note $b=1-a$, $c\gg a$, and $d\gg a$ for all experiments. (a) Loss $A$ vs $a$; (b) Loss $B$ vs $a$; (c) Loss $C$ vs $a$; (d) Loss $D$ vs $a$.  Note the small scale of the $y$-axis in (c) and (d); we see that $C\approx C_{opt}$ and $D\approx D_{opt}$ for all values of $a$.}
    \label{fig:validaionobjconf1}
\end{figure}

\textbf{Sensitivity to Choice of Hyperparameter $a$}\\
Recalling that $0 < a < 1$ represents the relative weight of the disparity in fairness loss $A$ compared to the regularization loss $B$, we repeat the above experiment for values of $a \in \{0.1, 0.2, 0.3, 0.4, 0.5, 0.6, 0.7, 0.8, 0.9, 0.99, 0.999, 0.9999\}$. We plot losses $A$, $B$, $C$, and $D$ for different values of $a$, averaging across 100 fits for each value of $a$, as shown in Figure~\ref{fig:validaionobjconf1}. 

As expected, when $A$ is given more importance in the total loss formulation, which occurs for large values of $a$, the value of $A$ is small (disparity is eliminated). When $B$ is given more importance in the total loss formulation, which occurs for small values of $a$, the value of $B$ is small (regularization loss is negligible). Note that the weight assigned to loss $B$ is $b=1-a$.  We observe that $C \approx C_{opt}$ and $D \approx D_{opt}$ for all values of $a$, since weights $c$ and $d$ are very large compared to $a$ and $b$.

\subsubsection{Validation of Theorem 4.3}

\textbf{Data Generation Process}\\
We generate a simple, synthetic dataset of size 100,000, split into 70,000 training records and 30,000 test records.  The dataset is generated to satisfy the preconditions of Theorem 4.3: there are no non-sensitive attributes $\mathbf{X}$, and the outcome $Y$ is conditionally independent of $S$ given the human decision $H$.  Moreover, there is $\delta$-unfairness toward $S=1$ in the observed decision $H$, i.e., $\text{logit}(H=1\:|\: S=1)-\mbox{logit}(H=1\:|\: S=0) = \delta$. We assume that $S$, $H$, and $Y$ are all binary.  We generate $S \sim \text{Bernoulli}(0.5)$. For the human decision, we generate $H \sim \text{Bernoulli}(0.01)$ for $S=0$, and $H \sim \text{Bernoulli}(0.6)$ for $S=1$.  For simplicity, we assume $Y=H$, i.e., the outcome is perfectly correlated with the human decision.

\begin{figure}[t]
\centering
\includegraphics[height=1.5cm,width=12cm]{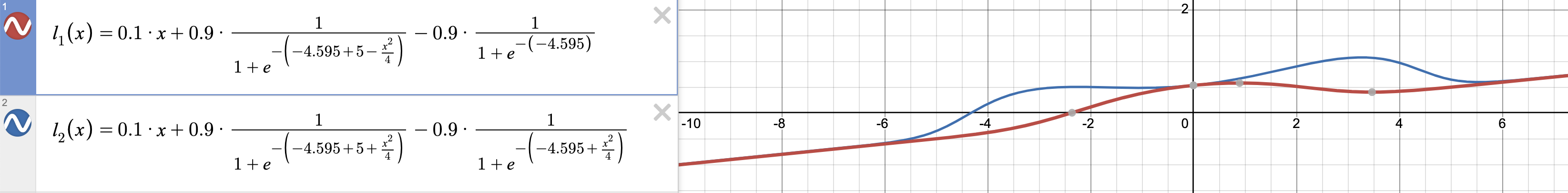}
\quad
\includegraphics[height=1.5cm,width=3cm]{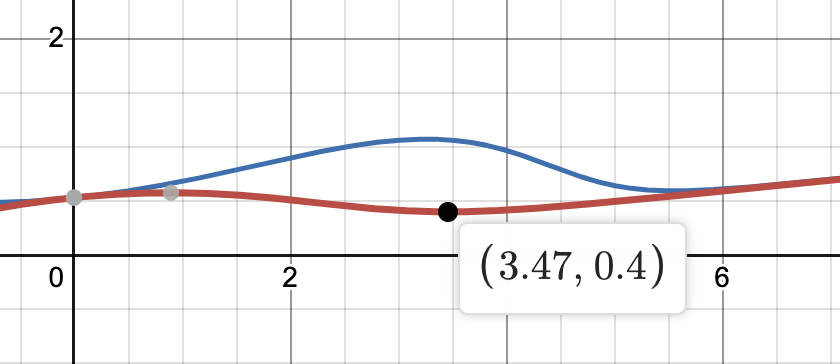}
\quad
\centering
\includegraphics[height=1.5cm,width=4cm]{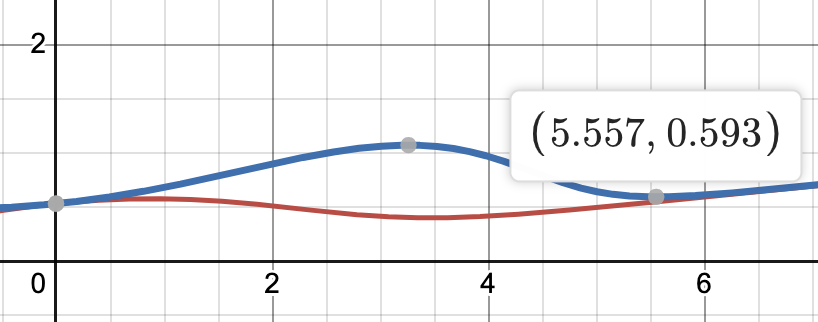}
\quad
\centering
\includegraphics[height=1.5cm,width=4cm]{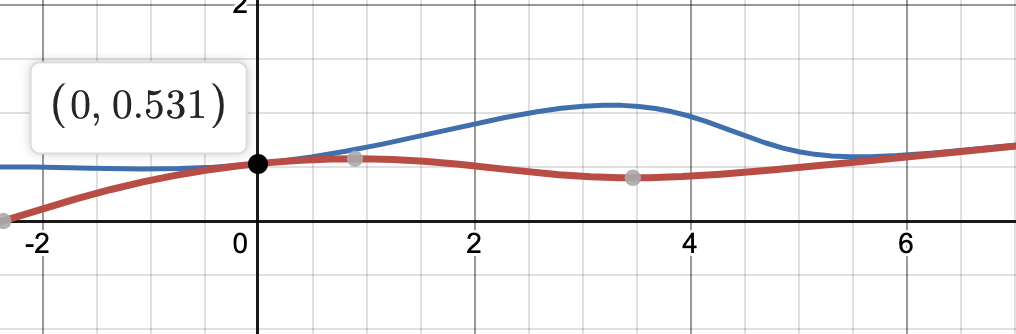}
\caption{Losses $L1$ and $L2$ (top), Case 1 (bottom left), 2 (bottom middle), and 3 (bottom right).  Loss $L1$ is shown in red, and Loss $L2$ is shown in blue. The $x$-axis is the regularization loss $B_\mathbf{w}=|w_3|+|w_{SR_3}|+|\text{bias}_{R_3}|$ for representational disparity node $R_3$, with corresponding shift in logits $B_\mathbf{w}^2/4$. The $y$-axis is total loss $aA_\mathbf{w} + bB_\mathbf{w}$ for $a=0.9$ and $b=0.1$.}\label{fig:lossl1}
\end{figure}

This corresponds to the conditions of Theorem 4.3 with parameters $\text{logit}(O(0))=-4.595$, $\delta=5$, and $\alpha=1$.  In this case, with weight of disparity loss $a=0.9$, we observe from Figure~\ref{fig:lossl1} that loss $L1$ (moving the larger logit toward the smaller logit, i.e., decreasing $\text{Pr}(H=1\:|\: S=1)$) is globally optimal.  In this scenario, Loss $L1 \approx 0.40$, and loss $L2 \approx 0.593$, where loss $L2$ would result from moving both logits to the extreme, i.e., increasing both  
$\text{Pr}(H=1\:|\: S=0)$ and $\text{Pr}(H=1\:|\: S=1)$.  However, we see below that, depending on the initialization, it is possible 
for the network to converge to the global optimum loss $L1$, the local optimum loss $L2$, or the ``no change'' loss $L0 \approx 0.531$, where the probabilities and therefore the disparity remain constant.

\textbf{Training}\\
We use the four-layer neural network architecture described in Section~\ref{ourmodel}.  The first layer consists of input $S$.  The second layer consists of three nodes $\mathbf{R}=\{R_1, R_2, R_3\}$ capturing the internal representation of $S$.  The third and fourth layers represent the human decision $H$ and outcome $Y$, as above.  The observed and desired decision-makers differ in the internal representations of the input used to make the human decision $H$. We assume that the observed decision-maker uses $\{R_1,R_2\}$, and the desired decision-maker uses $\{R_1, R_2, R_3\}$.  We use Adam optimizer~\cite{kingma2014adam} to minimize Eq.~\ref{totalloss1}, with hyperparameters $a=0.9$, $b=0.1$, $c=1000$, and $d=1000$.  (Note that Theorem 4.3 focuses on the case where $a$ and $b$ are similar in scale.)  We train the neural network on the training data for 1000 epochs for each fit, and average the results across 100 fits. 

\textbf{Effect of Initialization}\\
\ignore{We experimentally validated that each of the global minimum losses reported in Theorem 4.3 can be achieved by varying the initialization parameters. For illustration, we discuss in detail how a globally optimal loss L1 was achieved. Loss L1 is globally optimal when $a=0.9$, $\text{logit}(O(0))=-4.595$, $\delta=5$, and $\alpha=1$ as loss L1 is 0.4, loss L2 is 0.531, and when $\delta > 0$ the optimal loss is $\text{min}$(L1, L2). We randomly initialize all weights and find the optimal weights resulting in a loss close to the optimal loss L1. Below, we demonstrate the sensitivity of the result to the choice of initialization parameters. 

Note that the sign of initial weights being similar to the sign of optimal weights decides whether the converged loss is the global minimum loss, the local minimum loss, or no change loss.} We now demonstrate the importance of initialization by choosing initial parameters $w_3$, $w_{SR_3}$, and $\text{bias}_{R_3}$ for the representational disparity node $R_3$ that result in each of the three losses (global minimum $L1$, local minimum $L2$, or no change $L0$) discussed above.  For reproducibility, we provide all weights for each configuration in Table~\ref{tab:reproducibility}.

\textbf{Case 1:} When $w_3=-1.735$, $w_{SR_3}=1.735$, and $\text{bias}_{R_3}=0$ (same as the theoretically found optimal weights corresponding to loss $L1$), with other weights set to the weights resulting in the optimal loss $L1$, after training for $100$ iterations, the experimental loss obtained is $0.3996$, which is close to the global minimum loss of $0.4$.

\textbf{Case 2:} When $w_3=5$, $w_{SR_3}=0$, and $\text{bias}_{R_3}=5$ (similar to the theoretically found optimal weights corresponding to loss $L2$, only magnitude-scaled), with the other weights set to the weights resulting in the optimal loss $L1$, after training for $100$ iterations, the experimental loss obtained is $0.5982$, which is close to the local minimum loss of $0.593$ (loss $L2$). 

\textbf{Case 3:} When $w_3=-1$, $w_{SR_3}=1$, and $\text{bias}_{R_3}=-1$ (arbitrary weights not corresponding to loss $L1$ or loss $L2$), with the other weights set to the weights resulting in the optimal loss $L1$, after training for $100$ iterations, the experimental loss obtained is $0.5412$, which matches the no change loss of $0.531$ (initial loss $L0$). 
\begin{table}[t]
\centering
 \begin{tabular}{ | c | c | c | c |} 
 \hline
  & $R_1$ & $R_2$ & $R_3$ \\ [0.5ex] 
 \hline
 $S$ & 3.128 & -0.110 & 1.751 \\ 
 \hline
 \text{bias} & -0.046 & -0.327 & 0.052\\
 \hline
 \end{tabular}
\quad
 \begin{tabular}{ | c | c | } 
 \hline
  & $H$ \\ [0.5ex] 
 \hline
 $R_1$ & 1.645 \\
 \hline
 $R_2$ & 0.140 \\
 \hline
 $R_3$ & -1.770 \\
 \hline
 \text{bias} & -4.722 \\
 \hline
 \end{tabular}
%\end{table}
\quad
%\begin{table}[H]
%\caption{fc3 weights}
%\centering
  \begin{tabular}{ | c | c | } 
 \hline
  & $Y$ \\ [0.5ex] 
 \hline
 $R_1$ & 1.877 \\
 \hline
 $R_2$ & 38.032 \\
 \hline
 \text{bias} & -20.689\\
 \hline
 \end{tabular}
\quad
 \begin{tabular}{ | c | c | c | c |} 
 \hline
  & $R_1$ & $R_2$ & $R_3$ \\ [0.5ex] 
 \hline
 $S$ & 3.154 & -0.110 & 0 \\ 
 \hline
 \text{bias} & -0.022 & -0.327 & 2.854 \\
 \hline
 \end{tabular}
\quad
 \begin{tabular}{ | c | c | } 
 \hline
  & $H$ \\ [0.5ex] 
 \hline
 $R_1$ & 1.672 \\
 \hline
 $R_2$ & 0.140 \\
 \hline
 $R_3$ & 2.854 \\
 \hline
 \text{bias} & -4.691 \\
 \hline
 \end{tabular}
% \end{table}
  \quad
% \begin{table}[H]
% \caption{fc3 weights}
% \centering
 \begin{tabular}{ | c | c | } 
 \hline
  & $Y$ \\ [0.5ex] 
 \hline
 $R_1$ & 1.864 \\
 \hline
 $R_2$ & 38.026 \\
 \hline
 \text{bias} & -20.676 \\
 \hline
 \end{tabular}
 \quad
 \begin{tabular}{ | c | c | c | c |} 
 \hline
  & $R_1$ & $R_2$ & $R_3$ \\ [0.5ex] 
 \hline
 $S$ & 3.148 & -0.110 & -0.002 \\ 
 \hline
 \text{bias} & -0.038 & -0.327 & -0.003\\
 \hline
 \end{tabular}
\quad
 \begin{tabular}{ | c | c | } 
 \hline
  & $H$ \\ [0.5ex] 
 \hline
  $R_1$ & 1.654 \\
 \hline
 $R_2$ & 0.140 \\
 \hline
 $R_3$ & -0.009 \\
 \hline
 \text{bias} & -4.702 \\
 \hline
 \end{tabular}
%\end{table}
\quad
%\begin{table}[H]
%\caption{fc3 weights}
%\centering
  \begin{tabular}{ | c | c | } 
 \hline
  & $Y$ \\ [0.5ex] 
 \hline
 $R_1$ & 1.886 \\
 \hline
 $R_2$ & 38.013 \\
 \hline
 \text{bias} & -20.693 \\
 \hline
 \end{tabular}
\quad
 \caption{Learned fc1 (left), fc2 (middle), and fc3 (right) weights for Case 1 (top), 2 (middle), and 3 (bottom) respectively.}\label{tab:reproducibility}
\end{table}
\begin{table} [t]
\begin{center}
\begin{tabular}{p{1cm}|p{0.85cm}|p{0.85cm}|p{0.85cm}|p{1.2cm}|p{1.5cm}|p{0.85cm}|p{1.15cm}|p{1cm}}
\hline
& \textbf{Init. Disp.} & \textbf{A} & \textbf{B} & \textbf{aA + bB} &\textbf{Logit Shift}~$\left(\frac{B^2_\textbf{w}}{4}\right)$ & \textbf{C} & \textbf{D} & \textbf{Total Loss}   \\
\hline
\textbf{Case 1} &0.5915 &0.0471 &3.5722 &0.3996 &3.1899 &0.3601 &3.3E-09 &360.4836 \\ 
\textbf{Case 2} &0.5915 &0.0305 &5.7082 &0.5982 &8.1460 &0.3610 &3.4E-09 &361.6237 \\ 
\textbf{Case 3} &0.5915 &0.5999 &0.0132 &0.5412 &1.1E-16 &0.3600 &3.4E-09 &360.5110
\end{tabular} 
\caption{Comparison of losses when the network converges to the globally optimal loss $L1$ (Case 1), the locally optimal loss $L2$ (Case 2), or the ``no change'' loss $L0$ (Case 3).}
\label{compexpshifttodelta}
\end{center}
\end{table}

Table~\ref{compexpshifttodelta} lists the resulting losses for Cases 1 to 3. Let $A$, $B$, $C$, and $D$ be the training objectives discussed in Section~\ref{ourmodel} with total loss of $aA + bB + cC + dD$, where $a=0.9, b=0.1, c=1000$ and $d=1000$. Here, $A$ is the disparity loss, $B$ is the regularization loss, $C$ is the human decision cross-entropy loss, and $D$ is the outcome cross-entropy loss.

For Case 1, $aA + bB$ ($0.3996$) is close to the global minimum loss ($0.4$), however, the initial disparity of $0.5774$ is only reduced to $0.0471$ (not eliminated) as the logit decrease is $3.1899$, not equal to $\delta=5$ needed to eliminate disparity. For Case 2, $aA + bB$ ($0.5982$) is close to the local minimum loss of $0.593$ with logit in both sigmoids pushed to the extreme with a logit increase of $8.1460$. 
 For Case 3, $aA + bB$ ($0.5412$) is close to the initial loss of $0.531$ with negligible change in initial disparity and an insignificant logit change of $1.1\text{E-}16$. We observe that losses $C$ and $D$ remain close to their optimal values $C_{opt} \approx 0.360$ and $D_{opt} = 0$ for all cases.

Thus we observe that, when the weight of the disparity loss $a$ is not approximately equal to 1, some disparity remains even for the globally optimal solution.  Moreover, convergence to the globally optimal solution is not guaranteed, and whether the weights converge to the global minimum loss, local minimum loss, or no change loss depends on the initialization of weights $w$, $w_{SR'}$, and $\text{bias}_{R'}$ for the representational disparity node $R'$.  Thus we recommend performing multiple fits and choosing the one with lowest loss.

\subsection{Experiments on Real-World Data}
\label{expreal}

\noindent \textbf{Datasets} \\
Below we provide additional information on each of the three real-world datasets used in our experiments (Section~\ref{exp_sec}). 
\begin{enumerate}
\item \textbf{German Credit}: The dataset has 1,000 records. Each record has 20 attributes classifying account holders into a Good or Bad class. We consider $Age$ as the sensitive attribute, following~\cite{zemel2013learning, kamiran2009classifying}. We preprocess the data in the same manner as~\cite{zemel2013learning}, with 13 categorical attributes one-hot encoded and numerical attributes binarized at the median value. This resulted in 61 features excluding the target variable. 
    
\item \textbf{Adult income}: The dataset has 45,222 records. Each record has 14 attributes classifying whether or not an individual's income is larger than \$50,000. We consider $Gender$ as the sensitive attribute, following~\cite{zemel2013learning, kohavi1996scaling, kamishima2012fairness}. We preprocess the data in the same manner as~\cite{zemel2013learning} with 8 categorical attributes one-hot encoded and 6 numerical attributes binarized at the median value. This resulted in 103 features excluding the target variable. One-hot encoding results in some rows with $workclass$, $occupation$, and $native\text{-}country$ attributes with a missing value. We delete all rows with missing values.

\item \textbf{Heritage Health}: The dataset is from the Heritage Health Prize milestone challenge. We use features similar to the
winning team, Market Makers~\cite{marketmarkers}. The dataset has 184,308 records. The goal is to classify whether or not each individual will spend any days in the hospital that year. We run the SQL script in Appendix C of~\cite{marketmarkers} to generate records with features listed in Table Data Set 1. The $ageMISS$ feature denotes that the age value is missing, hence rows with $ageMISS=1$ are deleted, and the $ageMISS$ feature is dropped. Each categorical attribute was one-hot encoded, and each numerical attribute is binarized at the median value. Following~\cite{zemel2013learning}, we create a binary sensitive attribute $S$ whose value is 1 when age is older than 65 years and 0 otherwise. This translates to setting $S=0$ when $age\_05=1$ or $age\_15=1$ or $age\_25=1$ or $age\_35=1$ or $age\_45=1$ or $age\_55=1$ and $S=1$ otherwise. Preprocessing results in 143 binary features, excluding the target variable. Note that the preprocessed features (143 features) are not the same as~\cite{zemel2013learning} (1157 features) that uses features in both Table Data Set 1 and 2. In our work, our LRD approach and the competing LFR approach~\cite{zemel2013learning} are compared on the same dataset with 143 features. 
\end{enumerate} 

\textbf{Experiments}\\
As noted in Section~\ref{exp_sec}, we perform five different semi-synthetic experiments for each of the three real-world datasets, 
using the class variable as the human decision $H$.  Since these datasets do not have a downstream outcome that is separate from the class variable to be predicted, we generate a new outcome variable $Y$ which is dependent on $H$ and either adds to, partially mitigates, fully mitigates, or reverses the disparity (these are Cases I-IV respectively; Case V considers the special case where $Y=H$).  To create the distribution of $\text{Pr}(Y \:|\: H, S)$ for Cases I-IV, we first derive an expression for the total disparity 
$|\text{Pr}(Y=1\:|\:S=1)-\text{Pr}(Y=1\:|\:S=0)|$ in terms of our three experimental parameters,\begin{align*}
    a&=\text{Pr}(Y=1\:|\:H=1, S=s)-\text{Pr}(Y=1\:|\:H=0, S=s), ~\forall s \in \{0,1\},\\
    b&=\text{Pr}(Y=1\:|\:H=0, S=1)-\text{Pr}(Y=1\:|\:H=0, S=0), \text{ and} \\ c&=\text{Pr}(H=1\:|\:S=1)-\text{Pr}(H=1\:|\:S=0).
\end{align*}
We obtain:
\begin{align*}
&~~~~~|\text{Pr}(Y=1\:|\:S=1)-\text{Pr}(Y=1\:|\:S=0)| \\
&= |(\text{Pr}(Y=1\:|\:H=1,S=1) - \text{Pr}(Y=1\:|\:H=0,S=1))\text{Pr}(H=1\:|\:S=1) \nonumber \\ &~-(\text{Pr}(Y=1\:|\:H=1,S=0) - \text{Pr}(Y=1\:|\:H=0,S=0))\text{Pr}(H=1\:|\:S=0) \nonumber \\ &~+ \text{Pr}(Y=1\:|\:H=0,S=1) - \text{Pr}(Y=1\:|\:H=0,S=0)| \\ &=|ac+b|.
\end{align*}

Assuming a constant $a=0.6$ for Cases I-IV, and using the observed values of $c$ for each real-world dataset, we define $b=ac$ for Case I, $b=-0.5ac$ for Case II, $b=-ac$ for Case III, and $b=-1.5ac$ for Case IV.  For the Adult dataset, Case I, the value was clipped to -0.1 so that $\text{Pr}(Y=1 \:|\: S = 0, H = 1)$ does not exceed 1.
The resulting values are shown in Table~\ref{cbvalues}:
\begin{table} [t]
\begin{center}
\begin{tabular}{p{1.2cm}p{1.4cm}p{1.75cm}p{1.75cm}p{1.75cm}p{1.75cm}}
\hline
& \textbf{c} & \textbf{b (Case I)} & \textbf{b (Case II)} & \textbf{b (Case III)} & \textbf{b (Case IV)}   \\
\hline
German &~0.0953 &~0.0572 &-0.0286 &-0.0572 &-0.0858 \\ Adult &-0.1945 &-0.1000 &~0.0584 &~0.1167 &~0.1751 \\ Health &~0.0888 &~0.0533 &-0.0266 &-0.0533 &-0.0799
\end{tabular} 
\caption{Experimental setup. $c$ and $b$ values for Cases I-IV.}
\label{cbvalues}
\end{center}
\end{table}
In the German Credit and Health datasets, $H=1$ favors $S=1$, resulting in $c> 0$, and in the Adult dataset, $H=1$ favors $S=0$, resulting in $c<0$. The configuration tables for $\text{Pr}(Y=1\:|\: S=s,H=h)$ for Cases I-IV are given in Table~\ref{dispderi}; again, we note that Case V has $Y=H$.
\begin{table}[t]
\begin{center}
\begin{tabular}{ |p{0.9cm}|p{0.9cm}|p{2.25cm}| }
 \hline
 \text{$S=s$} & \text{$H=h$} & \text{$\text{Pr}(Y=1\:|\:s,h)$}\\
 \hline
 0 & 0 & 0.3\\
 \hline
 0 & 1 & 0.9\\
 \hline
 1 & 0 & $0.3+b$\\
 \hline
 1 & 1 & $0.9+b$\\
 \hline
\end{tabular}
\quad
\begin{tabular}{ |p{0.9cm}|p{0.9cm}|p{2.25cm}| }
 \hline
 \text{$S=s$} & \text{$H=h$} & \text{$\text{Pr}(Y=1\:|\:s,h)$}\\
 \hline
 0 & 0 & $0.3-b$\\
 \hline
 0 & 1 & $0.9-b$\\
 \hline
 1 & 0 & 0.3\\
 \hline
 1 & 1 & 0.9\\
 \hline
\end{tabular}
\end{center}
\caption{Experimental setup. Left: Configuration for German Credit and Health datasets. Right: Configuration for Adult dataset.}\label{dispderi}
\end{table}

\begin{figure}[t]
    \centering
    \subfigure[]{\includegraphics[width=4.5cm, height=3cm]{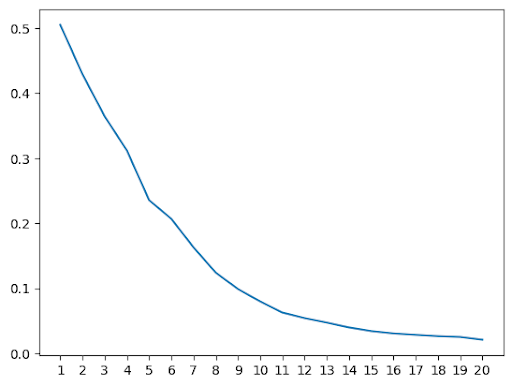}} 
    \subfigure[]{\includegraphics[width=4.5cm, height=3cm]{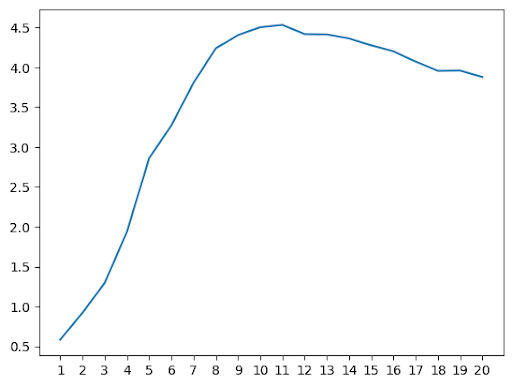}}
    \subfigure[]{\includegraphics[width=4.5cm, height=3cm]{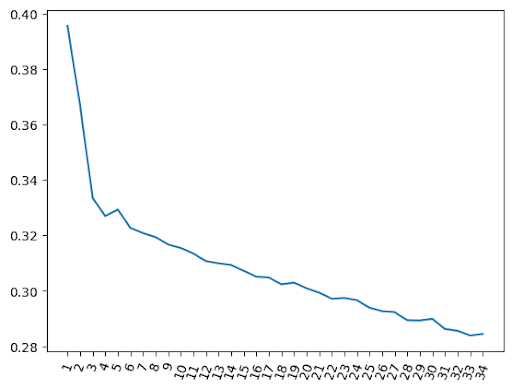}} 
    \subfigure[]{\includegraphics[width=4.5cm, height=3cm]{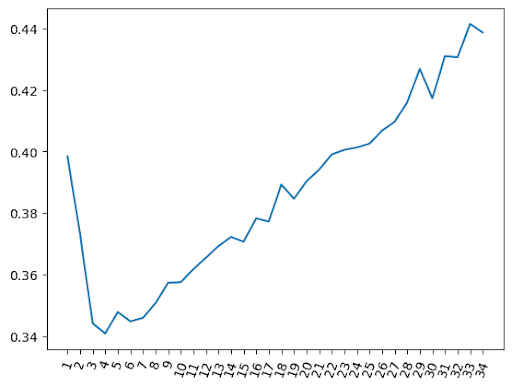}}
    \subfigure[]{\includegraphics[width=4.5cm, height=3cm]{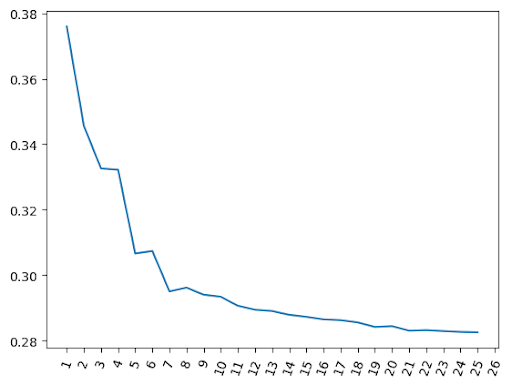}} 
    \subfigure[]{\includegraphics[width=4.5cm, height=3cm]{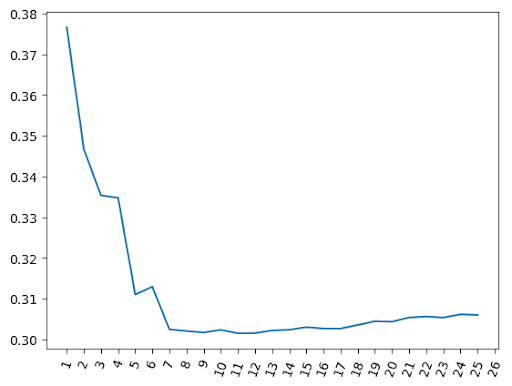}} 
    \caption{Selection of number of nodes $m'$ used to model the observed decision-maker by cross-validation.     
    (a) and (b) are the training and validation results respectively for the German Credit dataset. (c) and (d) are the training and validation results respectively for the Adult dataset. (e) and (f) are the training and validation results respectively for the Health dataset. For all graphs, the $x$-axis represents $m'$ and the $y$-axis represents cross-entropy loss.}
    \label{fig:modelselection}
\end{figure}

\textbf{Model Selection}\\
As noted in Section~\ref{exp_sec}, we use 5-fold cross-validation on the training data  to select the number of nodes used to model the observed decision-maker ($m'$). Total loss vs. $m'$ plots for model selection are shown in Figure~\ref{fig:modelselection}, and the number of nodes selected remains the same for Cases I to V. Based on the results, $m'=1$ for the German Credit dataset, $m'=4$ for the Adult dataset, and $m'=11$ for the Health dataset.

\textbf{Consistency}\\
In this section, we further elucidate consistency results. Let the correction from the observed human decision to the desired human decision be $\Delta \text{Pr}(H=1\:|\:\mathbf{X},S=s) = \text{Pr}_{\text{des}}(H=1\:|\:\mathbf{X},S=s) - \text{Pr}_{\text{obs}}(H=1\:|\:\mathbf{X},S=s)$. For Case V ($Y=H$) on the German Credit dataset, 5 of 10 splits resulted in $\Delta \text{Pr}(H=1\:|\:\mathbf{X},S=1)$ being reduced by at least 0.05 on average (range [-.15, 0]) while $\Delta \text{Pr}(H=1\:|\:\mathbf{X},S=0)$ stayed roughly the same (range [-.02, 0]). The other 5 splits resulted in $\Delta \text{Pr}(H=1\:|\:\mathbf{X},S=0)$ increasing by at least 0.05 on average (range [0, .20]) while $\Delta \text{Pr}(H=1\:|\:\mathbf{X},S=1)$ had a small increase of at most 0.07. We see similar results for Case V on the Adult and Health datasets, but with more consistency in the direction of correction. For the Adult dataset, 9 of 10 splits resulted in $\Delta \text{Pr}(H=1\:|\:\mathbf{X},S=0)$ being reduced by at least -.20 on average (range [-.21, 0]) and $\Delta \text{Pr}(H=1\:|\:\mathbf{X},S=1)$ was roughly the same (range [-0.01, 0]). For the Health dataset, all 10 splits resulted in $\Delta \text{Pr}(H=1\:|\:\mathbf{X},S=1)$ being reduced by -.09 on average (range [-.11, 0]) and $\Delta \text{Pr}(H=1\:|\:\mathbf{X},S=0)$ was roughly the same (range [-.02, 0]).  This demonstrates the high consistency of the LRD results, in contrast to the LFR method which had wide variation in individual probabilities: for LFR, many observations in each class have substantial increases and substantial decreases in probability, as measured by wide ranges of $\Delta \text{Pr}(H=1\:|\:\mathbf{X},S=s)$ for both $S=0$ and $S=1$. 

To compare LRD and LFR, we propose the following measures, 
\begin{align*}
\text{Consistency Measure (\textbf{CM})} &= \mathbb{E}_{d \sim D}\mathbb{E}_{s \sim d(S)}\text{Var}(\text{Pr}_\text{des}(H=1\:|\:\textbf{x},s) - \text{Pr}_\text{obs}(H=1\:|\:\textbf{x},s)\:|\:S=s), \\
\text{Consistency Ratio}~(\textbf{CR}) &= \frac{\text{\textbf{CM} for LFR}}{\text{\textbf{CM} for LRD}},
\end{align*}
where $d$ is a test split drawn from the dataset $D$, $s$ is sampled from values of $S$ in $d$, and $\text{Var}(\text{Pr}_\text{des}(H=1\:|\:\textbf{x},s) - \text{Pr}_\text{obs}(H=1\:|\:\textbf{x},s)\:|\:S=s)$ is the variance calculated across $x$ in $d$ with $S=s$. Table~\ref{consistencymeasure} reports the results for the German Credit, Adult, and Health test data averaged across 10 data splits. The results indicate that the average within-class variance of LRD's shifts from the observed probability to the desired probability is small, as indicated by \textbf{CM}, while the average within-class variance for LFR's shifts is substantially larger, as indicated by \textbf{CR}. 

\begin{table} [t]
\begin{center}
\begin{tabular}{@{}p{2.5cm} p{2cm}p{1.5cm}@{}}
\hline
& \textbf{CM} & \textbf{CR} \\
\hline
\textbf{Case I} ~~~German &0.0093 &21.59  \\ {~~~~~~~~~~~~~~~Adult} &0.0519 &3.06  \\ {~~~~~~~~~~~~~~~Health} &0.0133 &9.06 \\ \textbf{Case II} ~~German &0.0020 &100.4 \\ {~~~~~~~~~~~~~~~Adult} &0.0054 &29.44  \\ {~~~~~~~~~~~~~~~Health} &0.0007 &172.1 \\ \textbf{Case III} German &0.0008 &251.0 \\
{~~~~~~~~~~~~~~~Adult} &$1.92 \times 10^{-9}$ &$8.28 \times 10^7$\\
{~~~~~~~~~~~~~~~Health} &$2.58 \times 10^{-10}$ &$4.67 \times 10^8$\\ \textbf{Case IV} German &$8.91 \times 10^{-5}$ &$2.25 \times 10^3$ \\
{~~~~~~~~~~~~~~~Adult} &0.0055 &28.91 \\
{~~~~~~~~~~~~~~~Health} &0.0004 &301.3 \\ 
\textbf{Case V} ~~German &0.0031 &64.77 \\
{~~~~~~~~~~~~~~~Adult} &0.0263 &6.05 \\
{~~~~~~~~~~~~~~~Health} &0.0027 &44.63
\end{tabular} 
\caption{Consistency measure \textbf{CM} for LRD, and consistency ratio \textbf{CR} = \textbf{CM}(LFR) / \textbf{CM}(LRD).}
\label{consistencymeasure}
\end{center}
\end{table}

\end{document}